\documentclass[10pt,journal,compsoc]{IEEEtran}
\usepackage{xspace}
\newcommand{\ie}{i.e.\xspace}
\usepackage{graphicx}
\usepackage{booktabs}
\usepackage{amsmath}
\usepackage{colortbl}
\usepackage{array}
\usepackage{url}
\usepackage{xcolor}
\usepackage{array}
\usepackage{graphicx}
\usepackage{float} 
\usepackage{tcolorbox}
\usepackage{algorithm, algorithmicx, algpseudocode}
\usepackage{multirow}
\usepackage{amssymb}

\ifCLASSOPTIONcompsoc
  \usepackage[nocompress]{cite}
\else
  \usepackage{cite}
\fi
\ifCLASSINFOpdf
\else
\fi
\begin{document}
%
\title{Human Grasp Generation for Rigid and Deformable Objects with Decomposed VQ-VAE}
%
%
%
%

\author{Mengshi Qi,~\IEEEmembership{Member,~IEEE,}
        Zhe Zhao,
        Huadong Ma,~\IEEEmembership{Fellow,~IEEE} 
\thanks{This work is partly supported by the Funds for the NSFC Project under Grant 62202063, Beijing Natural Science Foundation (L243027). (\emph{Corresponding author: Mengshi Qi~(email:~qms@bupt.edu.cn)})}
\thanks{M. Qi, Z. Zhao, and H. Ma are with State Key Laboratory of Networking and Switching Technology, Beijing University of Posts and Telecommunications, China.}
}

\IEEEtitleabstractindextext{%
\begin{abstract}
Generating realistic human grasps is crucial yet challenging for object manipulation in computer graphics and robotics. Current methods often struggle to generate detailed and realistic grasps with full finger-object interaction, as they typically rely on encoding the entire hand and estimating both posture and position in a single step. Additionally, simulating object deformation during grasp generation is still difficult, as modeling such deformation requires capturing the comprehensive relationship among points of the object's surface. To address these limitations, we propose a novel improved Decomposed Vector-Quantized Variational Autoencoder~(DVQ-VAE-2), which decomposes the hand into distinct parts and encodes them separately. This part-aware architecture allows for more precise management of hand-object interactions. Furthermore, we introduce a dual-stage decoding strategy that first predicts the grasp type under skeletal constraints and then identifies the optimal grasp position, enhancing both the realism and adaptability of the model to unseen interactions. Furthermore, we introduce a new Mesh UFormer as the backbone network to extract the hierarchical structural representations from the mesh and propose a new normal vector-guided position encoding to simulate the hand-object deformation. In experiments, our model achieves a relative improvement of approximately \(14.1\%\) in grasp quality compared to state-of-the-art methods across four widely used benchmarks. Our comparisons with other backbone networks show relative improvements of \(2.23\%\) in Hand-object Contact Distance and \(5.86\%\) in Quality Index on deformable and rigid object based datasets, respectively. Our source code and model are available at \url{https://github.com/florasion/D-VQVAE}.

\end{abstract}

}

\maketitle 
\IEEEdisplaynontitleabstractindextext

%
\IEEEpeerreviewmaketitle

\IEEEraisesectionheading{\section{Introduction}\label{sec:introduction}}

%
%
%
%

\IEEEPARstart{G}{enerating} realistic human grasps for diverse objects~\cite{jiang2021hand,karunratanakul2020grasping,karunratanakul2021skeleton,zheng2023coop} is crucial in a variety of fields such as robotics, human-computer interaction, augmented reality and scene understanding~\cite{qi2019attentive,qi2021semantics,qi2020stc}. While substantial progress has been made in 3D hand pose estimation~\cite{ge2018hand,zimmermann2017learning,boukhayma20193d} and hand-object 3D reconstruction~\cite{hasson2019learning,fan2017point,tzionas20153d}, grasp generation remains a challenging task, requiring not only the accurate modeling of hand-object interactions but also a deeper understanding of the complex and sophisticated hand movements. The primary challenge in generating high-quality grasps lies in accurately modeling the interaction between the hand and objects, which can be divided into rigid and deformable categories. For rigid objects, once a grasp is generated, no further manipulation is necessary. However, deformable objects introduce an additional challenge~\cite{xie2023hmdo,xie2023nonrigid}, as the deformation induced by the hand’s interaction must be considered in the grasp generation process.

\begin{figure}[t]
    \centering
    \includegraphics[width=0.9\linewidth]{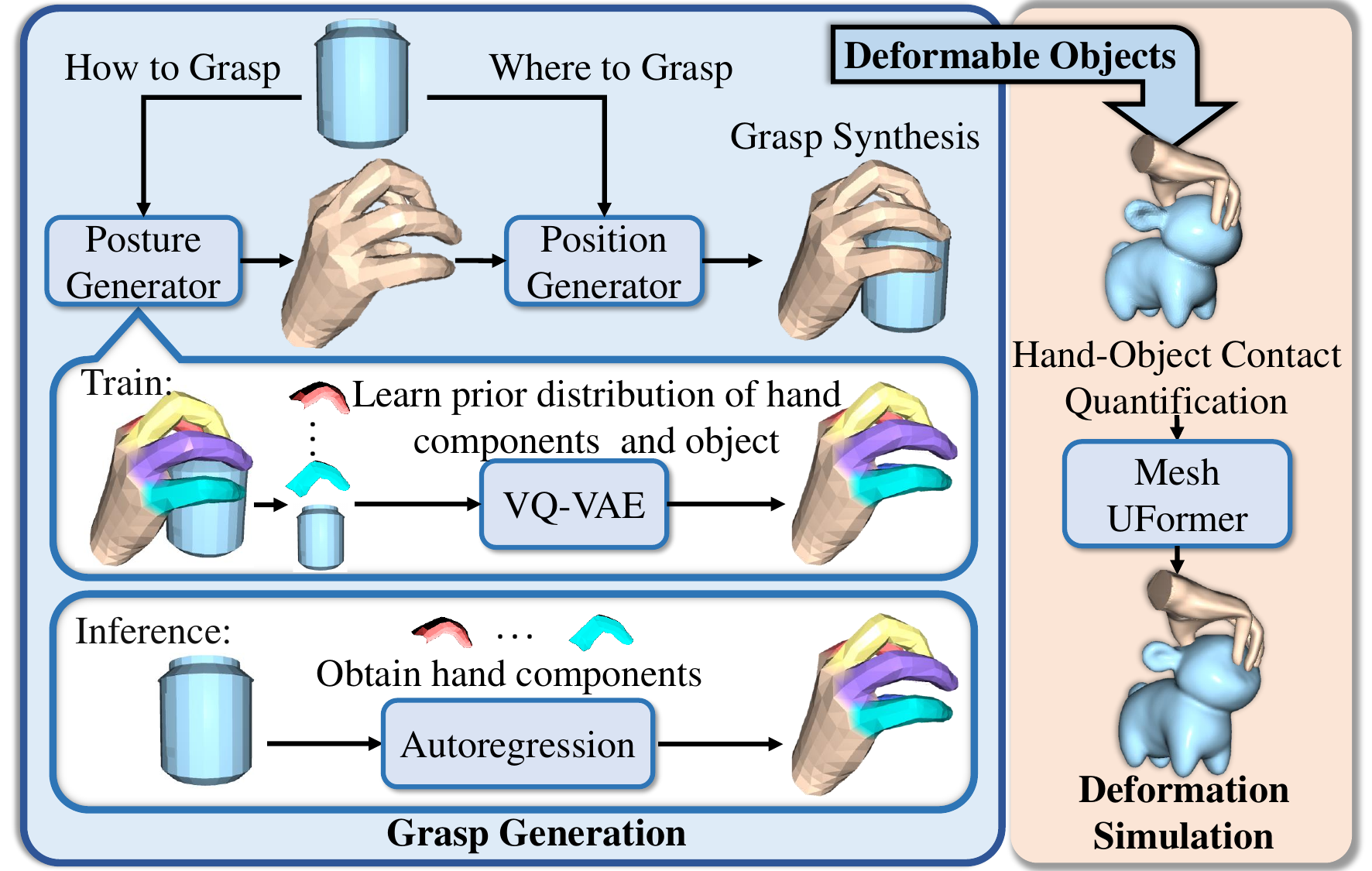}
    \vspace{-3mm}
    \caption{Illustration of our proposed grasp generation model. First, we employ Decomposed VQ-VAE-2~(DVQ-VAE-2) to learn the prior distributions of the object and each hand component~(\ie, five fingers and the palm) during training. Specifically, the decoding process is divided into two stages: generating hand posture and generating hand position. During inference, we use autoregression guided by the object to generate realistic human grasps. While for deformable objects, we propose a new Mesh UFormer to simulate deformations.}
    \label{fig:enter-label}
    \vspace{-3mm}
\end{figure}

Recent efforts in grasp generation~\cite{jiang2021hand,karunratanakul2020grasping,karunratanakul2021skeleton,zheng2023coop} for rigid objects have primarily focused on predicting contact maps to ensure full hand-object contact. While these methods improve grasp accuracy, they often result in unrealistic grasps that resemble touches rather than true grasps~\cite{liu2023contactgen}. Additionally, many of these approaches rely on Conditional Variational Autoencoders (CVAE)\cite{jiang2021hand,taheri2020grab,karunratanakul2021skeleton,karunratanakul2020grasping,liu2023contactgen,zheng2023coop}, with some incorporating Generative Adversarial Networks (GANs)\cite{corona2020ganhand,goodfellow2014generative}. However, these models often use a continuous latent space to represent all types of grasps, which fails to account for the discrete nature of human hand movements, leading to limited diversity and unrealistic grasps. In contrast, the Vector Quantized Variational Autoencoder (VQ-VAE)~\cite{van2017neural} offers a more controlled approach by utilizing discrete latent codes rather than a continuous latent space. This enables better grasp generation by capturing the discrete and categorical nature of hand movements. Inspired by such ideas, we adopt the VQ-VAE~\cite{van2017neural} as the core of our network, by extending it with a decomposed architecture that improves the utilization of individual hand components and increases diversity through autoregressive inference.

Furthermore, another key limitation of existing methods is their lack the adaptability to unseen or out-of-domain objects. Some studies have proposed test-time adaptation~\cite{jiang2021hand,karunratanakul2021skeleton}, which refines the model during inference by adjusting MANO parameters~\cite{romero2022embodied}. However, these approaches significantly increase inference time. To address this, we propose a two-stage grasping process: firstly predicting the hand pose based on the expected grasp type, and then determining the optimal position for manipulating the object.

For deformable objects, the interaction with the hand typically causes greater penetration than with rigid objects, due to their deformable nature. Current methods~\cite{jiang2021hand,karunratanakul2020grasping} designed for rigid-object with hand interaction fail to accurately model this contact with non-rigid objects. To overcome this, we introduce a new hand-object interaction quantification method. Moreover, simulating deformation of deformable objects requires a more profound understanding of their surface characteristics. Rather than relying on point clouds~\cite{qi2017pointnet,wu2024point}, we propose normal vector-guided position encoding and introduce a hierarchical Mesh UFormer network to integrate the structural information of the object's mesh and quantify the hand-object contact. 

In this work, to generate diverse and realistic human grasps for a given rigid or deformable object, we introduce a novel Decomposed VQ-VAE (DVQ-VAE), as shown in Fig.~\ref{fig:enter-label}. Our innovation lies in partitioning the hand into multiple components, each encoded in discrete latent spaces, followed by grasp generation using a dual-stage decoding strategy. Specifically, we propose a part-aware decomposed architecture that divides the hand into six distinct components: the five fingers and the palm, with each component encoded into its own codebook. Grasp generation occurs in two sequential steps by generating the hand's posture and then determining the optimal grasp position. More importantly, it should be mentioned that this paper is an extension of our conference paper~\cite{zhao2025decomposed}, and we improve the DVQ-VAE to DVQ-VAE-2. Compared to the original version, we incorporate interactions with deformable objects and simulate their deformations, by proposing a new backbone network, Mesh UFormer to integrate surface structural information from the object’s mesh. Furthermore, we design a novel hand-object interaction quantification method based on the normal vector-guided position encoding, to simulate the deformation by calculating the distance between hand and object mesh vertex. In experiments, we perform the deformable grasp generation on the new dataset, HMDO Dataset~\cite{xie2023hmdo}, and conduct more comprehensive tests, present additional visualizations, and provide a detailed ablation study to demonstrate the effectiveness of each component.

Our main contributions can be summarized as follows:

\par\textbf{(1)} We propose a novel DVQ-VAE for human grasp generation, utilizing a part-aware decomposed architecture to encode each hand component and progressively generate the entire hand. 

\par\textbf{(2)} We introduce a new dual-stage decoding strategy that improves the quality of generated grasps for unseen objects by gradually determining the grasp type under skeletal physical constraints and its position.

\par\textbf{(3)} We design a novel backbone network for processing meshes, \emph{i.e.,} Mesh UFormer to enhance the hierarchical mesh structure and representations.

\par\textbf{(4)} We present a deformation modeling approach for non-rigid objects, by adopting the normal vector-guided position encoding to quantify the hand-object interaction distance and improve the generation performance.

\par\textbf{(5)} We conduct extensive comparisons with other commonly-used methods on both deformable and rigid object datasets with five widely-adopted benchmarks, achieving relative improvements of \(2.23\%\) on Hand-object Contact Distance and \(5.86\%\) on Quality Index over the previous best results, respectively.

\section{Related Work}
\label{sec:formatting}


\noindent\textbf{Grasp Generation.} With advancements in virtual reality and robotics, substantial efforts have been made to generate physically plausible and diverse grasps, primarily using generative models such as GANs~\cite{goodfellow2014generative,corona2020ganhand} and VAEs~\cite{kingma2013auto,jiang2021hand,liu2023contactgen,karunratanakul2021skeleton,zheng2023coop}. A notable example is the work by Jiang \emph{et al.}\cite{jiang2021hand}, who introduced a Conditional VAE (CVAE)\cite{sohn2015learning} for grasp generation, incorporating objects as conditioning factors. This approach also employed test-time adaptation using ContactNet~\cite{jiang2021hand}. Later, Karunratanakul \emph{et al.}\cite{karunratanakul2021skeleton} proposed a keypoint-based skeleton articulation to improve the precision of hand grasping. Additionally, several works devoted to use reinforcement learning~\cite{zhang2025graspxl} and diffusion-based method~\cite{wang2024single}. Recently, VQ-VAE~\cite{van2017neural} has shown its effectiveness across various domains, such as image generation and 3D synthesis~\cite{mittal2022autosdf,pi2023hierarchical}. Inspired by the work of Pi \emph{et al.}\cite{pi2023hierarchical}, which segmented the human body into five parts and encoded them into multiple codebooks to generate human motion, we extend this concept by encoding the hand into multiple components. By utilizing VQ-VAE\cite{van2017neural}, we encode hand-object interaction data as discrete variables, allowing us to obtain a prior distribution of these interactions, which is then leveraged for grasp prediction. 

\noindent{\bf Hand-Object Interaction.} Modeling hand-object interaction~\cite{romero2022embodied,hasson2019learning,brahmbhatt2020contactpose,grady2021contactopt,tzionas2016capturing,wang2021interactive,yang2024learning,10478195,10599825} is a crucial task, yet the high degrees of freedom in hand poses present significant challenges. A pioneering work by Romero~\emph{et al.}\cite{romero2022embodied} introduced the parameterized hand model MANO to address the complexities of reconstructing hand pose and shape. Following this, Hasson \emph{et al.}\cite{hasson2019learning} introduced the large-scale synthetic dataset \emph{ObMan} and designed an end-to-end model for jointly reconstructing 3D hand-object interactions from RGB images. Brahmbhatt \emph{et al.}\cite{brahmbhatt2020contactpose} presented ContactPose, the first dataset focused on hand-object contact, and Grady \emph{et al.}\cite{grady2021contactopt} proposed ContactOpt to improve the quality of hand-object contact modeling. Liu~\emph{et al.}\cite{10599825} proposed the HCTransformer to address the domain shift in human-object action recognition. However, these approaches assume that both the hand and the object are visible in the input. In contrast, our approach predicts a realistic grasp for the observed rigid or deformable objects without requiring visibility of the hand.

\begin{figure*}[t]
    \centering
    \includegraphics[width=0.9\linewidth]{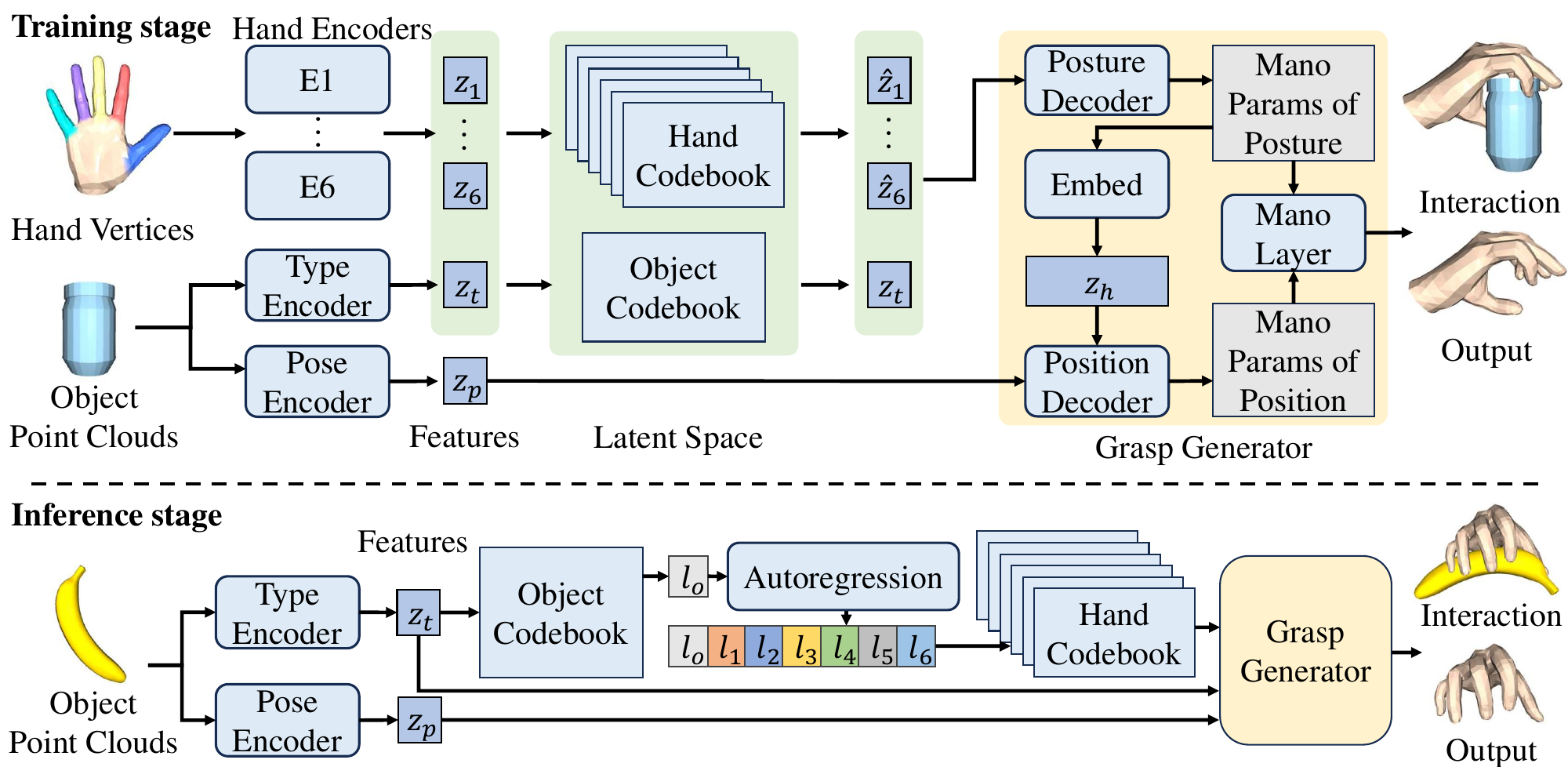}
    \vspace{-2mm}
    \caption{Overall architecture of the proposed DVQ-VAE model, which follows the encoder-decoder paradigm. During training, the model takes hand vertices and object point clouds as inputs, and maps them to discrete latent spaces comprising seven codebooks~(\ie, one for the object and six for different hand components) to generate hands capable of grasping objects. During inference, only object point clouds are used as input to generate hands grasping the given object.}
    \vspace{-2mm}
    \label{fig:network}
\end{figure*}

\noindent{\bf Deformable Objects Simulation.} Traditional hand-object interaction models are designed to model contact between the hand and rigid objects~\cite{jiang2021hand,liu2023contactgen,brahmbhatt2020contactpose,grady2021contactopt,tzionas2016capturing,karunratanakul2020grasping}. In these cases, the hand does not penetrate deeply into the object, and the contact region on the hand is typically fixed, serving as a prior. However, when interacting with deformable objects, the hand often penetrates significantly into the object to induce deformation. In such cases, traditional methods are no longer applicable, and the hand priors become ineffective. So far, research on the object's deformation during hand-object interactions is rather limited, although significant progress~\cite{xie2023hmdo,xie2023nonrigid,zhang2019interactionfusion} has been made in the reconstruction of deformable objects. Most of these methods employ the Deformation Graph~\cite{sumner2007embedded} based approach, by sampling a subset of points on the deformable object and modeling the entire object mesh as a graph. The nodes of this graph are then manipulated, and the remaining points are adjusted by minimizing a target function, which primarily serves to ensure the surface of the deformed object remains smooth. This approach is effective for simulating deformations caused by shape editing, such as turning a doll's head to the left or lifting its hand. However, in order to accurately simulate hand-object contact deformation, it is essential to consider every contact point on the object. Sampling only a subset of points can result in the loss of critical contact information. In contrast, our proposed model \emph{Mesh UFormer} takes the features of all object points as input and outputs the displacement for each point during deformation modeling, followed by a unified smoothing process.


\noindent{\bf Object Modeling.} Traditional object modeling in hand-object interaction~\cite{jiang2021hand,brahmbhatt2020contactpose,karunratanakul2020grasping,liu2023contactgen} typically encode only object point features through PointNet~\cite{qi2017pointnet} or PointTransformer~\cite{wu2024point}. Yet, these methods separate the points from the mesh, relying solely on the distances between points for their relationships, without utilizing the mesh information. Leveraging mesh information to enhance the model's understanding of object shape is crucial for improving the quality of hand-object interaction modeling. In mesh processing, methods like MeshMAE~\cite{liang2022meshmae} and Mesh U-Nets~\cite{beetz2022mesh}, downsample the mesh and gradually refine it into patches. However, these approaches can disrupt the original mesh structure. To address this, we propose a new backbone model Mesh UFormer that utilizes mesh information while preserving the original mesh structure. We voxelized the input mesh, clustering mesh points within each voxel into a single point. Additionally, we establish point mapping between the new voxelized mesh and the original mesh to maintain the original structure and then improve the simulated deformation performance. 

\section{ Grasp Generation for Rigid Objects}
\label{sec:appro}

For rigid objects, our objective is to create a stable, physically plausible hand mesh for grasping by utilizing an object point cloud as input. To accomplish this, we introduce a decomposed VQ-VAE, structuring the decoding process into two sequential stages to produce the final grasp configurations, as shown in Fig.~\ref{fig:network}. 

\subsection{Overview}
\label{subsec:arch}

{\bf Training.} As depicted in the upper part of Fig.~\ref{fig:network}, our model takes the sampled object points \( P^o \in \mathbb{R}^{N_o \times 3} \) and hand vertices \( P^h \in \mathbb{R}^{778 \times 3} \) as inputs, using two encoders to extract object type and pose features (refer to \( z_t \) and \( z_p \)), respectively. Simultaneously, the hand vertices are divided into \( N = 6 \) segments, comprising the five individual fingers (thumb, index, middle, ring, and little finger) and the palm, as shown in Fig.~\ref{fig:network}. Each segment is separately encoded, yielding hand features \( z_f = \{z_1, \ldots, z_N\} \). These hand features, along with \( z_t \), are then mapped into a discrete latent space where \( N + 1 \) distinct codebooks \( Z = \{Z_o, Z_1, \ldots, Z_N\} \) are learned. During decoding, we employ MANO~\cite{romero2022embodied} to generate the hand mesh. MANO uses several parameter sets: \( \alpha \in \mathbb{R}^{10} \) for person-specific hand shape, \( \beta \in \mathbb{R}^{45} \) for joint rotations, \( \gamma \in \mathbb{R}^{3} \) for hand translation, and \( \delta \in \mathbb{R}^{3} \) for hand rotation. Here, \( \alpha \) and \( \beta \), related to hand posture, are combined as \(\hat{M}_{\text{posture}}\), while \( \gamma \) and \( \delta \), linked to hand position, are referred to as \(\hat{M}_{\text{position}}\). The closest matching vectors \(\hat{z}_f = \{\hat{z}_1, \ldots, \hat{z}_N\}\) are concatenated with \( z_t \) and passed into the Posture Decoder to decode \(\hat{M}_{\text{posture}}\). The encoded \(\hat{M}_{\text{posture}}\) (denoted \( z_h \)) is then combined with \( z_p \) and input to the Position Decoder to produce \(\hat{M}_{\text{position}}\). Finally, both parameter sets pass through the MANO layer to generate the final hand mesh.

{\bf Inference.} As shown in the bottom part of Fig.~\ref{fig:network}, the model takes sampled object points \( P^o \in \mathbb{R}^{N_o \times 3} \) as input, obtaining object features \( z_t \) and \( z_p \) through two object encoders. Next, \( z_t \) is matched with its nearest vector in the object codebook \( Z_o \) to determine an index \( l_o \). An autoregressive model, conditioned on \( l_o \), is then used to produce a sequence of hand codebook indices \( l_h = \{l_1, \ldots, l_N\} \). These indices retrieve corresponding hand features \(\hat{z}_f = \{\hat{z}_1, \ldots, \hat{z}_N\}\) from the hand codebooks. Lastly, these hand features, along with \( z_t \) and \( z_p \), are input into the grasp generator to create the desired hand mesh.

In this model, PointNet and our proposed Mesh UFormer described in Sec.~\ref{sec:MeshUFormer} can serve as the encoder for both object and hand point clouds, and all decoders are multi-layer perceptrons (MLPs). For inference, PixelCNN~\cite{van2016conditional} is used as the autoregressive model.


\subsection{Object Encoder}
\label{subsec:dis}


Unlike existing methods~\cite{jiang2021hand,karunratanakul2020grasping,karunratanakul2021skeleton,liu2023contactgen,zheng2023coop,taheri2020grab}, we introduce two distinct object encoders: a type encoder and a pose encoder. The type encoder operates on the premise that grasp types are diverse and intrinsically related to the shape of the object. By extracting \( z_t \), we facilitate the learning of an object codebook in the latent space, capturing these grasp types and categorizing them into common groups, enabling the model to determine suitable grasps for any input object. In contrast, the pose encoder is designed to capture features crucial for decoding grasp locations, specifying where the hand should make contact with the object. This division of the object encoder into two components allows the object codebook to remain independent of object positioning, thereby enriching the model with complementary features.


\subsection{Part-Aware Decomposed Architecture}
\label{sec:Part-Aware Decomposed Architecture}

Here we describe the proposed part-aware decomposed architecture of DVQ-VAE. Traditional VQ-VAEs~\cite{van2017neural} are typically used for image generation and employ a single codebook. This stems from the same codebook indices can be located in various positions of the image, while during autoregression the same codebook also can be used to infer the pixel at any position. However, in the case of human hands, the positions of different fingers are fixed. Therefore, we propose to extend VQ-VAE~\cite{van2017neural} to a part-aware decomposed architecture, \ie, encoding the object and the N parts of the hand as \(z_t\) and \(z_f\), respectively, and search for the vectors with the minimum Euclidean distance in their respective discrete codebooks during the latent embedding. To prevent the loss of object features, we use the discovered \(\hat{z}_f\) in combination with the original \(z_t\) as input for the grasp generator. In this scenario, the object's latent variable serves solely as a condition during autoregression. Thus, the training of the object codebook is conducted through unsupervised learning. In this context, to encourage \(z_t\) and \(z_f\) to closely approximate the matched vectors in the codebooks, the loss function \(\mathcal{L}_E\) for codebook training can be defined as follows:
\begin{equation}
\mathcal{L} _{h}= \sum_{i=1}^{N} \left \| sg(\hat{z}_i )- z_i \right \|^2_2+\beta  \sum_{i=1}^{N} \left \| sg( z_i)- \hat{z}_i\right \|^2_2,
\end{equation}
\begin{equation}
    \mathcal{L} _{o}=  \left \| sg(\hat{z}_t )- z_t \right \|^2_2+\beta   \left \| sg( z_t)- \hat{z}_t\right \|^2_2,
\end{equation}
\begin{equation}
    \mathcal{L}_E=\lambda_e\cdot(\mathcal{L}_h+\mathcal{L}_o),
\end{equation}
where
\begin{equation}
    \hat{z}_i = e_k,~\text{where}~k=argmin_j\left\|z_{i} - e_j\right\|_2,
    \label{eq1}
\end{equation}
\begin{equation}
    \hat{z}_t= e_m,~\text{where}~m=argmin_j\left\|z_t - e_j\right\|_2,
\end{equation}
where the operator \(sg(\cdot)\) denotes the halting of gradient flow, thereby preventing gradients from propagating into the associated parameter. Here, \(\lambda_e\) is a hyperparameter, and \(e\) represents an embedding in the codebook.

During inference, following the approach of VQ-VAE~\cite{van2017neural}, we employ PixelCNN~\cite{van2016conditional} as our autoregressive model. The object's codebook indices serve both as the condition and the initial sequence for PixelCNN~\cite{van2016conditional}, which predicts a sequence of codebook indices corresponding to the hand components.



\begin{figure*}[t]
    \centering
    \includegraphics[width=0.95\linewidth]{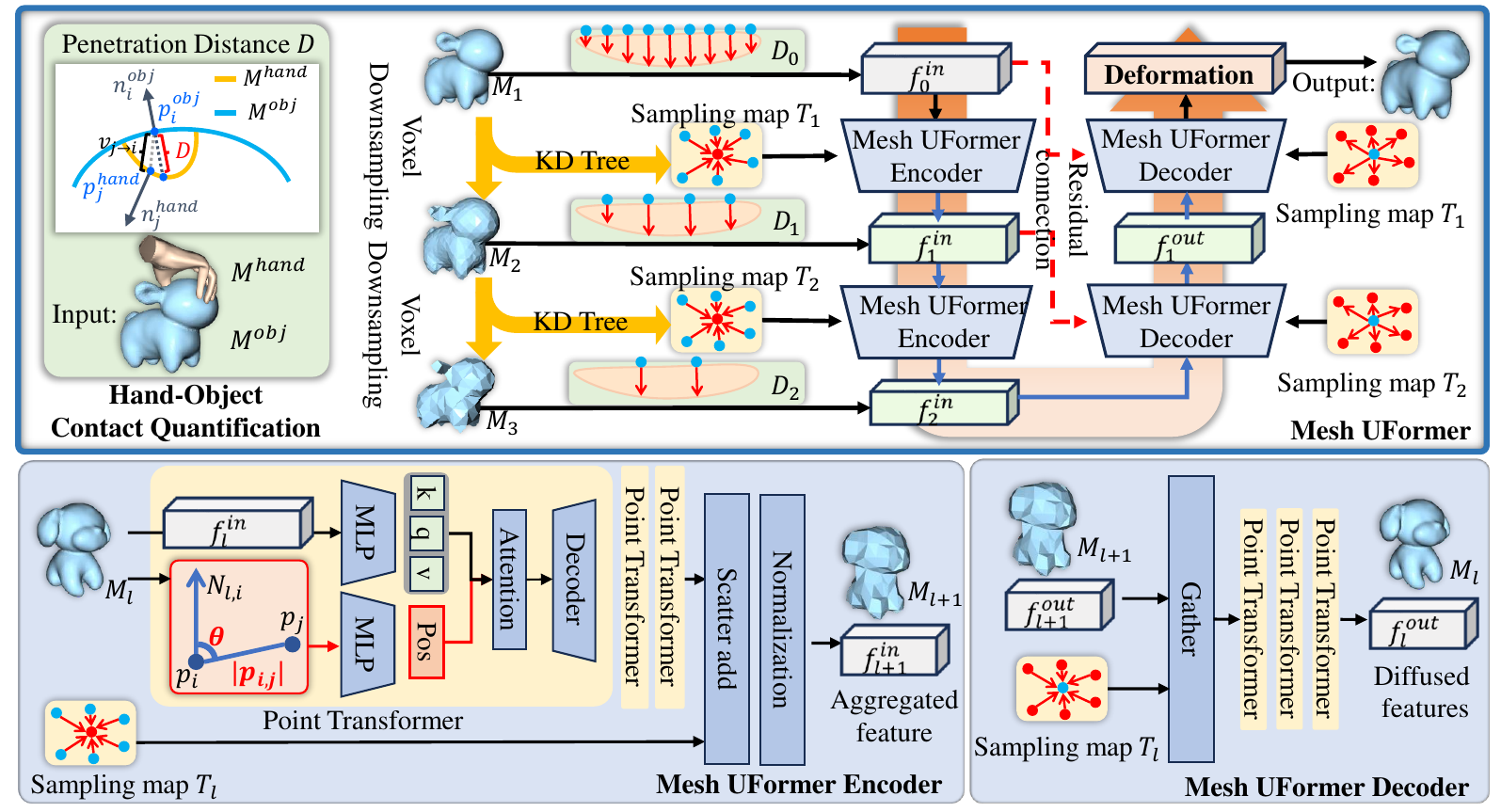}
    \vspace{-3mm}
    \caption{The upper part of the figure illustrates the overall architecture of our DVQ-VAE-2 with Mesh UFormer used for deformation simulation. We employ our newly proposed hand-object contact quantification method to construct the input for each point on the object, combined with voxel-based down-sampling to create a hierarchical input. Our Mesh UFormer is then used to simulate the deformations. The lower part of the figure details the structure of the encoder and decoder in the Mesh UFormer. These modules use a simplified PointTransformer integrated with our proposed normal vector-guided position encoding, combined with Scatter Add and Gather operations to achieve feature forward mapping and backward mapping between layers.}
    \vspace{-3mm}
    \label{fig:network_2}
\end{figure*}

\subsection{Dual-Stage Decoding Strategy}
\label{sec:Dual-Stage Decoding Strategy}


Previous methods~\cite{jiang2021hand,liu2023contactgen} enhanced grasp quality by generating 61 MANO~\cite{romero2022embodied} parameters in a single step and optimizing these parameters across multiple iterations, which led to considerable time consumption. To address the limitations of single-step decoding, we propose a dual-stage decoding strategy that sequentially generates the grasp posture and grasp position. In line with this decoupling approach, we separate the MANO parameters into two parts: position and posture. We first generate the more complex grasp posture parameters \(\hat{M}_{Posture}\) and then produce the fewer position parameters \(\hat{M}_{position}\) by integrating hand features with object characteristics.


1) To facilitate the generation of realistic grasps during decoding, we compute the L2 distance between the ground truth \(M_{Posture}\) and predicted values as follows:
\begin{equation}
\mathcal{L}_{posture} =\left \|M_{posture}-\hat{M}_{posture}  \right \|_2.
\end{equation}
In addition, we apply physical constraints informed by hand-skeletal dynamics. By extracting skeletal key points \(J\) from the reconstructed hand, we calculate angles between adjacent key points. These angles enable the creation of gating information to refine grasps that deviate from natural hand-skeletal configurations:
\begin{equation}
\theta_i = \cos^{-1} \left( \frac{\overrightarrow{J_iJ_{i-1}} \cdot \overrightarrow{J_iJ_{i+1}}}{\|\overrightarrow{J_iJ_{i-1}}\| \|\overrightarrow{J_iJ_{i+1}}\|} \right),
\end{equation}
\begin{equation}
\hat{M}_{posture}=\hat{M}_{posture}+G(\theta) \odot T(\hat{M}_{posture}),
\end{equation}
where \(\theta\) denotes the joint angles, \(G(\cdot)\) is the network generating gating information, and \(T(\cdot)\) is the transformer layer that provides correction values.

2) For training the position decoder, we prevent gradient propagation at \(z_h\) and calculate the L2 distance between \(M_{position}\) and hand vertices as:
\begin{equation}
\hat{M}_{position}=Dec[sg(z_h),z_p],
\end{equation}
\begin{equation}
\mathcal{L}_{position} =\left \|M_{position}-\hat{M}_{position}  \right \|_2,
\end{equation}
\begin{equation}
\mathcal{L}_{v} =\left \|P^h-M(\hat{M}_{position},\hat{M}_{posture})  \right \|_2,
\end{equation}
where \(Dec[\cdot]\) represents decoding through the position decoder, and \(M(\cdot)\) refers to passing through the MANO layer~\cite{romero2022embodied}. 

The reconstruction loss \(L_R\) combines \(\mathcal{L}_{posture}\), \(\mathcal{L}_{position}\), and \(\mathcal{L}_{v}\) as:
\begin{equation}
\mathcal{L}_{R} =\lambda_h\cdot(\mathcal{L}_{posture}+\mathcal{L}_{position})+\lambda_v\cdot\mathcal{L}_{v},
\end{equation}
where \(\lambda\) are hyperparameters. To ensure the model does not inadvertently learn hand position details prior to the position decoder, we center the hand vertices by subtracting their mean coordinates during training.

\subsection{Optimization}
\label{sec:Loss Function}

Finally, our total objective loss comprises three components: \(\mathcal{L}_E\) for the discrete latent embeddings described in Sec.~\ref{sec:Part-Aware Decomposed Architecture}, \(\mathcal{L}_R\) for constraining the morphology of the generated hand detailed in Sec.~\ref{sec:Dual-Stage Decoding Strategy}, and \(\mathcal{L}_{contact}\) for ensuring appropriate contact between the hand and the manipulated object, which we describe below.

Following the approach in~\cite{jiang2021hand}, we utilize object-centric contact loss \(\mathcal{L}_c\) and contact map consistency loss \(\mathcal{L}_m\) to improve the interaction between the hand and the object, formulated as:

\begin{equation}
    \mathcal{L}_c=\sum_{p_m\in P_m }^{} \min_{p_c\in P_c}\left | p_m-p_c \right |  ,
\end{equation}
\begin{equation}
    \mathcal{L}_m=\frac{\left |P_m  \right | \cap \left | \hat{P}_m  \right | }{\left |P_m  \right |},
\end{equation}

\noindent where the operator \(\left |\cdot \right |\) denotes the number of points in the set. \(P_m\) and \(\hat{P}_m\) are the point sets of the ground truth and the predicted grasp contact map, respectively, while \(P_c\) represents the points on the hand that could potentially contact the object. \(p_c\) and \(p_m\) refer to individual points in the sets \(P_c\) and \(P_m\), respectively.

We also incorporate penetration loss to enhance the realism of the generated grasp, formulated as:
\begin{equation}
\mathcal{L}_p=  {\textstyle \sum_{p\in P_{in} }\left \| p-P^o_i \right \|^2_2  } ,
\end{equation}
where
\begin{equation}
P^o_i=\text{argmin}_{p_i\in{P^o} } \left \| p-p_i \right \|,
\end{equation}
with \(P_{in}\) being the subset of hand points that enter the object. Here, \(p\) and \(p_i\) denote points in the sets \(P_{in}\) and \(P^o\), respectively. Thus, the contact loss can be expressed as:

\begin{equation}
    \mathcal{L}_{contact}=\lambda_m\cdot\mathcal{L}_{m}+\lambda_c\cdot\mathcal{L}_{c} +\lambda_p\cdot\mathcal{L}_p,
\end{equation}
where \(\lambda\) are hyperparameters.

Finally, we combine \(\mathcal{L}_{contact}\) with \(\mathcal{L}_E\) and \(\mathcal{L}_R\) to formulate our loss function for proposed DVQ-VAE:
\begin{equation}
\mathcal{L}_{} =\mathcal{L}_{R}+\mathcal{L}_{E}+\mathcal{L}_{contact}.
\label{eq2}
\end{equation}

\section{Grasp Generation for Deformable Objects}

For deformable objects, we improve the DVQ-VAE to DVQ-VAE-2 by introducing a new backbone named Mesh UFormer instead of PointNet. Furthermore, we propose a new hand-object deformation modeling approach to quantify the hand-object contact and then simulate the deformation, as shown in Fig.~\ref{fig:network_2}.

\subsection{Mesh UFormer}
\label{sec:MeshUFormer}

Since PointNet~\cite{qi2017pointnet} can only process point cloud input yet overlook the structural information of the whole mesh, we propose a new backbone to improve our DVQ-VAE, named Mesh UFormer. Inspired by the original U-Net architecture~\cite{ronneberger2015u}, our proposed Mesh UFormer adopts a symmetric encoder-decoder structure, utilizing 3D meshes as both inputs and outputs. To enhance information flow between the early and later layers of the network, we integrate skip connections linking corresponding levels of the encoder and decoder branches.

Specifically, given an input mesh $M$ and the feature vector \(\mathbf{f}\) of each point on the corresponding mesh, our model is capable of predicting an output \(\mathbf{\hat{f}}\) for each point \(i\). To prepare inputs for the Mesh UFormer, we construct a sequence of voxel grids with increasing sizes, denoted as \(\mathcal{V}=\{\mathcal{V}_1, \mathcal{V}_2, \dots, \mathcal{V}_L\}\), where each \(\mathcal{V}_l\) represents the voxel size at layer \(l\). For each voxel size \(\mathcal{V}_l\), we apply voxel clustering to sample the mesh, resulting in a sequence of meshes \({M}=\{{M}_1, {M}_2, \dots, {M}_L\}\). Moreover, we utilize the KD Tree~\cite{moore1990efficient} to perform efficient spatial partitioning and fast nearest neighbor searches, and then we compute point-to-point mappings between adjacent meshes based on their proximity, yielding a mapping sequence \({T}=\{{T}_1, {T}_2, \dots, {T}_{L-1}\}\), where \({T}_l\) represents the mapping from \({M}_l\) to \({M}_{l+1}\). Based on the length \(L\) of the sequence, we design the Mesh UFormer with \(L-1\) layers. Each layer in the Mesh UFormer passes features between meshes \({M}_l\) and \({M}_{l+1}\), maintaining the mesh structure. Each layer consists of two main components: a Mesh UFormer encoder for forward mapping features, and a Mesh UFormer decoder for backward mapping features.

\textbf{Mesh UFormer Encoder:} Each Mesh UFormer encoder takes the output features \(\mathbf{f}^{in}_{l-1}\) from the previous layer (or the original features \(\mathbf{f}_0\)), the normal vector \(\mathcal{N}_l\) of each point extracted from the mesh \({M}_l\), and the mapping \({T}_l\) as inputs. The input features are processed via a simplified Point Transformer as follows:

Computing \(k, q, v\): Queries \(q_i\), keys \(k_i\), and values \(v_i\) are computed using multi-layer perceptrons (MLPs):
\begin{equation}
   q_i,k_i,v_i = \text{MLP}(\mathbf{f}^{in}_{l-1}(i)).
\end{equation}
   
Calculating pairwise distances: For each point \(p_i\), computing the distance \(\|p_{ij}\|\) between \(p_i\) and other points \(p_j\), and then selecting \(K\)-nearest neighbors:
\begin{equation}
    \|{p}_{ij}\| = \|{p}_i - {p}_j\|.
\end{equation}

Gathering  neighbors: Selecting the \(K\)-nearest neighbors and computing the normal vector-guided position encodings \(\mathbf{e}_{ij}\):
\begin{equation}
\theta_{ij} = \arccos \left( \frac{{p}_{ij} \cdot \mathcal{N}_{l,i}}{\|{p}_{ij}\| \|\mathcal{N}_{l,i}\|} \right),
\end{equation}
\begin{equation}
\mathbf{e}_{ij} = \text{PosEnc}\left(\|\mathbf{p}_{ij}\|,\theta_{ij}\right),
\end{equation}
where \(\mathcal{N}_{l,i}\) is the normal vector at point \({p}_i\), \(\theta_{ij}\) is the angle between vectors \({p}_{ij}\) and \(\mathcal{N}_{l,i}\), and \(\text{PosEnc}(\cdot)\) is implemented by an MLP.

Computing attention weights: Based on the queries \(q_i\), keys \(k_j\), and positional encodings \(\mathbf{e}_{ij}\), calculating the attention weights \(\alpha_{ij}\):
\begin{equation}
   \alpha_{ij} = \operatorname{dropout}\left(\operatorname{softmax}\left(\frac{q_i \cdot \left(k_j + \mathbf{e}_{ij}\right)^\top}{\sqrt{\operatorname{dim}_{k_j}}}\right)\right).
\end{equation}

Applying the attention: Utilizing the attention weights to aggregate the value features \(v_j\) and then apply a linear transformation to the aggregated features to produce the output \(\mathbf{f}^{in'}_{l}\):
\begin{equation}
   \mathbf{f}^{in'}_{l}(i) = \text{MLP}(\sum_{j \in \mathcal{N}(i)} \alpha_{ij} v_j),
\end{equation}

\noindent where $\mathcal{N}$ denotes the neighbour set of point $i$. Then the processed features \(\mathbf{f}^{in'}_l\) can be scattered to the next mesh based on the mapping \(\mathcal{T}_l\) followed by normalization, yielding the aggregated output features \(\mathbf{f}^{in}_{l+1}\):
\begin{equation}
    \mathbf{f}^{in}_{l+1}(j) = 
    \frac{1}{|\mathcal{T}_l(j)|} \sum_{i \in \mathcal{T}_l(j)} \mathbf{f}^{in'}_l(i).
\end{equation}

\textbf{Mesh UFormer Decoder:} The Mesh UFormer Decoder propagates input features \(\mathbf{f}^{out'}_{l-1}\) to a denser mesh \(\mathcal{M}_l\) based on the mapping \(\mathcal{T}_l\).
\begin{equation}
    \mathbf{f}^{out'}_{l-1}(i) = \mathbf{f}^{out}_{l-1}(\mathcal{T}_l(i)).
\end{equation}

The backward mapping features \(\mathbf{f}^{out'}_{l-1}\) are combined with the input features from the corresponding Mesh UFormer Encoder via residual connections. This result is processed by the Point Transformer, producing the final output features \(\mathbf{f}^{out}_{l}\) as the following:
\begin{equation}
\mathbf{f}^{out}_{l} = \text{PointTransformer}\left(\mathbf{f}^{out'}_{l-1} + \mathbf{f}^{in}_{l}\right).
\end{equation}

During the forward mapping and backward mapping process, the point-to-point mappings \(\mathcal{T}_l\) ensure that the network preserves the mesh structure while integrating structural information into the point features.

\begin{figure}
    \centering
    \includegraphics[width=1\linewidth]{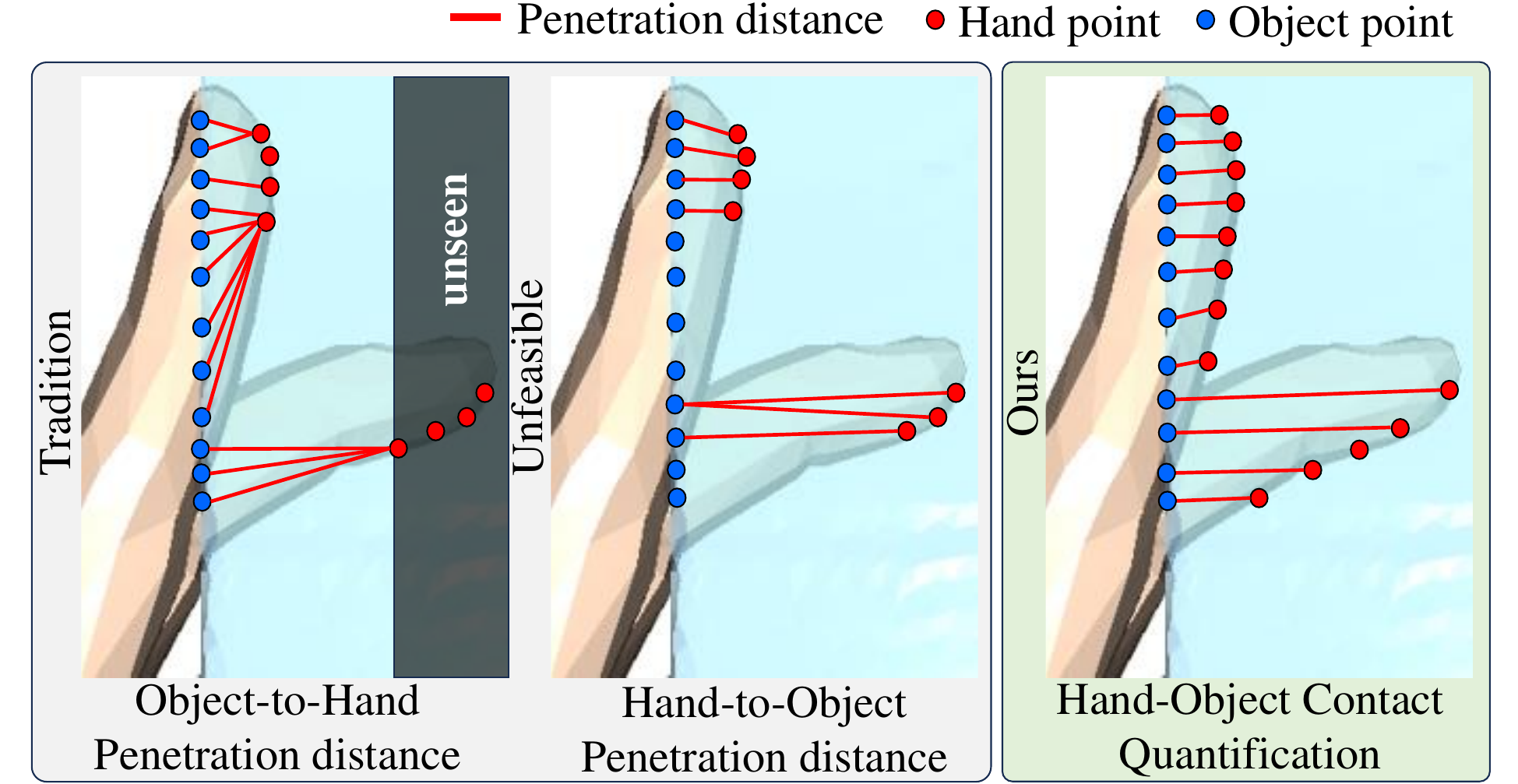}
    \vspace{-5mm}
    \caption{Comparison between our proposed Hand-Object Contact Quantification method and traditional methods. Unlike simply using the hand-to-object distance or the object-to-hand distance, we utilize both distances simultaneously.}
    \vspace{-3mm}
    \label{fig:contact}
\end{figure}


\subsection{Hand-Object Deformation Quantification}
\label{HoC}

As shown in Fig.~\ref{fig:contact}, traditional hand-object contact modeling methods are typically used to model the contact between rigid objects and the hand. These methods use fixed hand contact priors by pre-defining a set of hand contact points $\mathbf{P}^\text{hand-prior}$. For each object point $\mathbf{p}_i^\text{object}$, the closest hand contact prior point $\mathbf{p}_j^\text{hand-prior}$ is identified and then computing the distance $\|\mathbf{p}_i^\text{object} - \mathbf{p}_j^\text{hand-prior}\|$ for contact modeling. However, this approach may lead to the exclusion of hand-prior points that are too far away, and cause the object's contact map to fail to represent the true penetration distance of the hand into the object. In order to better model the interaction between hands and deformable objects, we propose a new hand-object contact quantification method considering the object and hand normal vector positions, as shown in Fig.~\ref{fig:network_2}.

Formally, given the mesh of hand ${M}^\text{hand}$ and the mesh of a deformable object ${M}^\text{obj}$, where the object mesh includes vertex sets ${\mathbf{p}_i^\text{object}}$ and their corresponding normal vectors ${\mathbf{n}_i^\text{object}}$. Our goal is to obtain a deformed object mesh ${M}^\text{deformed}$ by modeling hand-object deformation that conforms to the real-world scenario.

For each hand vertex $\mathbf{p}_j^\text{hand}$, the closest object point $\mathbf{p}_i^\text{object}$ can be identified, and the vector can be achieved as $\mathbf{v}_{j \rightarrow i} = \mathbf{p}_i^\text{object} - \mathbf{p}_j^\text{hand}$. Next, we define a sign value as \(d_{ji}\) by performing the dot product between vector \(\mathbf{v}_{j \rightarrow i}\) and the hand normal vector $\mathbf{n}_j^\text{hand}$:
\begin{equation}
    d_{ji} = \mathbf{v}_{j \rightarrow i} \cdot \mathbf{n}_j^\text{hand}.
\end{equation}

If $d_{ji} < 0$, it indicates that the hand point $\mathbf{p}_j^\text{hand}$ has penetrated the object mesh ${M}^\text{obj}$. All hand points that penetrate the object mesh are recorded as contact points.

For each object point $\mathbf{p}_i^\text{object}$, the closest contact point on the hand $\mathbf{p}_j^\text{hand}$ can be achieved, and the vector from $\mathbf{p}_i^\text{object}$ to $\mathbf{p}_j^\text{hand}$ is calculated as $\mathbf{v}_{i \rightarrow j}$. The dot product \(d_{ij}\) between vector \(\mathbf{v}_{i \rightarrow j}\) and the object normal vector $\mathbf{n}_i^\text{object}$ is then computed as following:
\begin{equation}
    d_{ij} = \mathbf{v}_{i \rightarrow j} \cdot \mathbf{n}_i^\text{object}.
\end{equation}

If $d_{ij} < 0$, it shows that the object point $\mathbf{p}_i^\text{object}$ is penetrated by a hand point $\mathbf{p}_j^\text{hand}$. Then, a ray is cast to calculate the penetration distance $D_i$ in the opposite direction of the object's normal vector, which is from the marked penetrated object point $\mathbf{p}_i^\text{object}$ to the hand mesh ${M}^\text{hand}$, as shown in the top left part of Fig.~\ref{fig:network_2}.

\subsection{Optimization}
\label{sec:Deformations loss Function}

To model the deformation of non-rigid objects, we use the Mesh UFormer described in Sec.~\ref{sec:MeshUFormer}. The input consists of a sequence of object meshes and the corresponding hand mesh, denoted as \( M^{\text{obj}} = \{ M^{\text{obj}}_1, M^{\text{obj}}_2, \ldots, M^{\text{obj}}_L \} \) and \( M^{\text{hand}} \). For each object mesh in the sequence, we apply the hand-object interaction modeling method outlined in Sec.~\ref{HoC} to compute the hand-object penetration distance as \( D = \{D_1, D_2, \ldots, D_L\} \), where $L$ is the length of the sequence. This distance serves as the input for the Mesh UFormer, allowing it to simulate the deformations \( \Delta \mathbf{p}^\text{object} \) as the following:
\begin{equation}
    \Delta \mathbf{p}^\text{object} = \text{Mesh UFormer}(M^{\text{obj}},D),
\end{equation}

\noindent where \( \Delta \mathbf{p}^\text{object} \in \mathbb{R}^{N \times 4} \) includes both the direction of deformation \( \mathbf{d}^\text{dir} \in \mathbb{R}^{N \times 3} \) and the magnitude of the deformation \(\Delta d \in \mathbb{R}^{N \times 1} \), where $N$ signifies the number of mesh vertices. For each vertex, we need to multiply these components and then yield the final deformation $\Delta \mathbf{p}^\text{object}$:
\begin{equation}
    \Delta \mathbf{p}^\text{object} = \mathbf{d}^\text{dir} \cdot \Delta d.
\end{equation}

The deformation displacement is applied to the original object mesh, resulting in the preliminary deformed object mesh ${M}^\text{deformed}$:
\begin{equation}
    {M}^\text{deformed} = M^\text{obj} + \Delta \mathbf{p}^\text{object}.
\end{equation}

To eliminate non-smoothness, the Taubin smoothing method~\cite{taubin1995curve} is applied to obtain the final smoothed deformed object mesh $M^\text{smooth}$ as following:
\begin{equation}
    {M}^\text{smooth} = \text{Taubin smoothing}({M}^\text{deformed}).
\end{equation}

During this process, we also calculate the laplacian smoothing loss~\cite{nealen2006laplacian} and normal consistency loss based on PyTorch3D~\cite{ravi2020accelerating} to constrain the smoothness of the mesh surface as follows:
\begin{equation}
\mathcal{L}_\text{laplacian}=\text{laplacian smoothing}({M}^\text{deformed}),
\end{equation}
\begin{equation}
\mathcal{L}_\text{normal}=\text{normal consistency}({M}^\text{deformed}).
\end{equation}

Furthermore, the mean squared error (MSE) loss between displacement and ground truth is adopted for each mesh vertex as the following formulation:
\begin{equation}
    \mathcal{L}_\text{MSE} = \frac{1}{N} \sum_{i=1}^N \|\Delta \mathbf{p}_i^\text{object} - \Delta \mathbf{p}_i^\text{gt}\|^2.
\end{equation}

Additionally, we achieve the point cloud similarity between the smoothed deformed object mesh and the ground truth via using chamfer distance from PyTorch3D~\cite{ravi2020accelerating}\footnote{\url{https://github.com/facebookresearch/pytorch3d}} as following:
\begin{equation}
\mathcal{L}_\text{similarity}=\text{chamfer distance}({M}^\text{deformed},{M}^\text{gt}).
\end{equation}

Finally, for simulation of hand-object deformation, our overall loss function \(\mathcal{L}_\text{Deform}\) is composed of the combination of the above-mentioned losses as follows:
\begin{equation}
\mathcal{L}_\text{Deform} =\mathcal{L}_\text{laplacian}+\mathcal{L}_\text{normal}+\mathcal{L}_\text{MSE}+\mathcal{L}_\text{similarity}.
\end{equation}

\section{Experiments}
\label{sec:exp}


\subsection{Datasets}
\label{subsec:dataset}


\noindent{\bf Obman Dataset~\cite{hasson2019learning}} is a comprehensive collection of synthesized manipulation interaction data produced by GraspIt~\cite{miller2004graspit}. It includes a vast array of 150,000 grasps across 2,772 distinct objects. Following the approach in~\cite{liu2023contactgen,jiang2021hand}, we designate 141,550 of these grasps for training and the remaining 6,285 for testing.

\noindent{\bf HO-3D~\cite{hampali2020honnotate}, FPHA~\cite{garcia2018first}, and GRAB~\cite{taheri2020grab} Datasets.} In line with~\cite{jiang2021hand}, we opted to train on the Obman dataset~\cite{hasson2019learning} due to its extensive variety of objects while testing on these three datasets. The test data serves as out-of-domain objects to evaluate adaptability. Specifically, HO-3D~\cite{hampali2020honnotate} contains 10 objects, FPHA~\cite{garcia2018first} includes 4 objects, and GRAB~\cite{taheri2020grab} comprises 51 objects collected from 10 subjects.

\noindent{\bf HMDO Dataset~\cite{xie2023hmdo}} contains 13 sequences of 12 deformable objects, comprising 15,000 instances of hand interactions with deformable objects. In experiments, we used this dataset to train our proposed deformation simulation network.

\subsection{Metrics}
\subsubsection{Grasp Generation for Rigid Objects}
\label{subsubsec:Rigid}



For a fair comparison, in accordance with~\cite{liu2023contactgen,jiang2021hand,karunratanakul2020grasping,karunratanakul2021skeleton,hasson2019learning,tzionas2016capturing}, we evaluate the results using the following metrics, where the \emph{quality index} is our proposed new metric. 

\noindent{\bf 1) Contact Ratio~($\%$).} This metric calculates the proportion of grasps that successfully contact the given object among all generated grasps.

\noindent{\bf 2) Hand-Object Interpenetration Volume~($cm^3$).} This is the volume shared by the 3D voxelized models of the hand and the object. Following~\cite{liu2023contactgen,karunratanakul2021skeleton}, we voxelize both the hand and object meshes using a size of 0.1 $cm^3$. 

\noindent{\bf 3) Grasp Simulation Displacement~(Grasp Disp)~($cm$).} This metric quantifies the stability of the generated grasp under simulated gravity, by reporting the average displacement of the object's center of mass.

\noindent{\bf 4) Entropy and Cluster Size.} Following the approach in~\cite{karunratanakul2021skeleton,liu2023contactgen}, we divide the generated grasps into 20 clusters using K-means, and measure both entropy and average cluster size to assess diversity. Higher entropy values and larger cluster sizes indicate the greater diversity.

\noindent{\bf 5) Time (\(s\)).} This metric evaluates the inference speed of the model in generating one batch of grasps.

\noindent{\bf 6) Quality Index.}~It is important to note that low penetration or low physical simulation displacement does not necessarily indicate a high-quality grasp. For instance, insufficient contact can lower penetration but may also compromise grasp stability, while severe penetration of the hand into the object can lead to reduced displacement in the physical simulator. Inspired by \cite{wang2024adaptive}, we use a utility function to quantitatively assess grasp quality:
\begin{equation}
Q =  a\cdot x+(1-a)\cdot y ,
\end{equation}
where \(a\) represents the weight used to balance the trade-off between grasp displacement (\(y\)) and penetration volume (\(x\)). According to \cite{wang2024adaptive}, we measure both displacement and penetration for each grasp in the Obman and GRAB datasets. We calculate \(a=0.301\) using the same method described in \cite{wang2024adaptive}. More details about the quality index please refer to our supplementary materials.

\subsubsection{Grasp Generation for Deformable Objects}

To evaluate the accuracy and performance of the model in simulating deformation during grasping, we used the following metrics:

\noindent{\bf 1) Hand-Object Contact Distance.} Following~HODistance~\cite{cao2021reconstructing}, this measure evaluates the accuracy of deformation in the expected regions of a deformable object, by calculating the distance within hand-object contact areas. The lower the value of this metric, the closer the distance of the contact, indicating accurate generating deformation. 

\noindent{\bf 2) Chamfer Distance.} Following \cite{ye2023diffusion}, we compute the Chamfer distance between the non-contact regions of the object and the original undeformed mesh. Specifically, we used the ground truth contact map as a mask to measure the Chamfer distance in areas not affected by contact. A lower value indicates less deformation in incorrect regions.

\noindent{\bf 3) Laplacian Smoothing Objective.}  Using methods from prior research~\cite{nealen2006laplacian,desbrun1999implicit}, we assessed the smoothness of the deformed mesh.  

\noindent{\bf 4) Contact Ratio.} As discussed in Sec.~\ref{subsubsec:Rigid}, we calculated the hand-object contact ratio.

\noindent{\bf 5) Number of Parameters.}  This metric reflects the model's parameter size.

\subsection{Compared Methods}
\label{subsec:baseline}

For rigid objects, we compare our proposed VQ-VAE with the following state-of-the-art methods: 
1) {\bf GraspTTA~\cite{jiang2021hand}}: This method is trained on the Obman dataset~\cite{hasson2019learning} and employs CVAE~\cite{sohn2015learning} to generate grasps conditioned on the object. It further refines the generated grasps using ContactNet with test-time adaptation (TTA). 
2) {\bf GraspCVAE~\cite{jiang2021hand}}: A variant of GraspTTA that does not utilize TTA. 
3) {\bf ContactGen~\cite{liu2023contactgen}}: Trained on the Grab dataset~\cite{taheri2020grab}, this method uses CVAE~\cite{sohn2015learning} to generate contact maps for the object, subsequently optimizing hand parameters based on these maps. 
4) {\bf Grasping Field~\cite{karunratanakul2020grasping}}: This method is trained on the Obman dataset~\cite{hasson2019learning} and utilizes VAE~\cite{kingma2013auto} to generate grasps based on the signed distance fields of the object and hand. 

For deformable objects, we compare our proposed method with the widely used deformation simulation method {\bf 1) Deformation Graph~\cite{sumner2007embedded}}: Optimizing global deformation based on the deformation of individual points. In addition, we compared our proposed Mesh UFormer with the following widely used point cloud processing methods: 2)~{\bf PointTransformerV3~\cite{wu2024point}}: A method that leverages self-attention mechanisms to capture local and global geometric features in point clouds. 3)~{\bf PointNet~\cite{qi2017pointnet}}: A framework that utilizes a symmetric function to aggregate features from unordered point sets, effectively handling permutation invariance in point cloud data. 4)~{\bf Standard Position Encoding}: Encoding the difference between the absolute coordinates of the target point and the center point. 5)~{\bf Object-to-Hand Contact Modeling}: As described in ~Sec.~\ref{HoC}, The hand-object contact is modeled by calculating the distance from the points on the hand to the nearest object point.


\begin{table*}[t!]
    \centering
    \caption{Performance comparison between the proposed DVQ-VAE-2 and state-of-the-art methods for grasp generation w.r.t rigid objects on widely-adopted benchmarks, \ie, HO-3D~\cite{hampali2020honnotate}, FPHA~\cite{garcia2018first}, GRAB~\cite{taheri2020grab}, and Obman~\cite{hasson2019learning}.} 
    \vspace{-3mm}
\begin{tabular}{@{}c|c|c|cccc|cc|c@{}}
\toprule
Dataset & Method & \begin{tabular}[c]{@{}l@{}}Object\\ Encoder\end{tabular} & \begin{tabular}[c]{@{}c@{}}Contact \\ ratio~\((\%)\)↑\end{tabular} & \begin{tabular}[c]{@{}c@{}}Penetration\\ Volume \((cm^3)\)↓\end{tabular} & \begin{tabular}[c]{@{}c@{}}Grasp \\ Disp \((cm)\) ↓\end{tabular} & \begin{tabular}[c]{@{}c@{}}Time\\ ~\((s)\) ↓\end{tabular} & \begin{tabular}[c]{@{}c@{}}Entr-\\ opy~↑\end{tabular} & \begin{tabular}[c]{@{}c@{}}Cluster \\ Size~↑\end{tabular} & \begin{tabular}[c]{@{}c@{}}Quality\\ Index↓\end{tabular} \\ \midrule
\multicolumn{1}{c|}{\multirow{6}{*}{\begin{tabular}[c]{@{}c@{}}HO-3D\\ ~\cite{hampali2020honnotate}\end{tabular}}} & \multicolumn{1}{c|}{GraspCVAE~\cite{jiang2021hand}} & \multicolumn{1}{c|}{PointNet} & \underline{99.60} & 7.23 & 2.78 & \multicolumn{1}{c|}{\textbf{0.0040}} & \textbf{2.96} & \multicolumn{1}{c|}{0.81} & \underline{4.12} \\
\multicolumn{1}{c|}{} & \multicolumn{1}{c|}{GraspTTA~\cite{jiang2021hand}} & \multicolumn{1}{c|}{PointNet} & \textbf{100} & 9.00 & \underline{2.65} & \multicolumn{1}{c|}{19.67} & 2.87 & \multicolumn{1}{c|}{0.80} & 4.56 \\
\multicolumn{1}{c|}{} & \multicolumn{1}{c|}{ContactGen~\cite{liu2023contactgen}} & \multicolumn{1}{c|}{PointNet++} & 90.10 & 6.53 & 3.72 & \multicolumn{1}{c|}{119.4} & \underline{2.94} & \multicolumn{1}{c|}{\textbf{4.79}} & 4.57 \\
\multicolumn{1}{c|}{} & \multicolumn{1}{c|}{GraspingField~\cite{karunratanakul2020grasping}} & \multicolumn{1}{c|}{PointNet} & 89.60 & 20.05 & 4.14 & \multicolumn{1}{c|}{57.49} & 2.91 & \multicolumn{1}{c|}{3.31} & 8.93 \\
\multicolumn{1}{c|}{} & \multicolumn{1}{c|}{DVQ-VAE} & \multicolumn{1}{c|}{PointNet} & 99.50 & \underline{5.36} & 2.75 & \multicolumn{1}{c|}{0.14} & 2.80 & \multicolumn{1}{c|}{3.84} & \underline{3.54} \\
\multicolumn{1}{c|}{} & \multicolumn{1}{c|}{\cellcolor[HTML]{E0E0E0}DVQ-VAE-2} & \multicolumn{1}{c|}{\cellcolor[HTML]{E0E0E0} Mesh UFormer} & \cellcolor[HTML]{E0E0E0} 99.00 & \cellcolor[HTML]{E0E0E0}\textbf{4.39} & \cellcolor[HTML]{E0E0E0}\textbf{2.33 } & \multicolumn{1}{c|}{\cellcolor[HTML]{E0E0E0}\underline{0.11}} & \cellcolor[HTML]{E0E0E0}2.82 & \multicolumn{1}{c|}{\cellcolor[HTML]{E0E0E0}\underline{3.86}} & \cellcolor[HTML]{E0E0E0}\textbf{2.95} \\ \midrule
\multicolumn{1}{c|}{\multirow{5}{*}{\begin{tabular}[c]{@{}c@{}}FPHA \\ ~\cite{garcia2018first}\end{tabular}}} & \multicolumn{1}{c|}{GraspCVAE~\cite{jiang2021hand}} & \multicolumn{1}{c|}{PointNet} & 98.98 & 7.46 & 2.97 & \multicolumn{1}{c|}{\textbf{0.0038}} & \underline{2.91} & \multicolumn{1}{c|}{0.81} & 4.32 \\
\multicolumn{1}{c|}{} & \multicolumn{1}{c|}{GraspTTA~\cite{jiang2021hand}} & \multicolumn{1}{c|}{PointNet} & \textbf{100} & 8.26 & \textbf{2.75} & \multicolumn{1}{c|}{19.29} & \textbf{2.93} & \multicolumn{1}{c|}{0.76} & 4.41 \\
\multicolumn{1}{c|}{} & \multicolumn{1}{c|}{ContactGen~\cite{liu2023contactgen}} & \multicolumn{1}{c|}{PointNet++} & 94.00 & 10.43 & 3.64 & \multicolumn{1}{c|}{237.38} & 2.84 & \multicolumn{1}{c|}{2.88} & 5.68 \\
\multicolumn{1}{c|}{} & \multicolumn{1}{c|}{GraspingField~\cite{karunratanakul2020grasping}} & \multicolumn{1}{c|}{PointNet} & 97.00 & 29.78 & 5.47 & \multicolumn{1}{c|}{58.57} & 2.85 & \multicolumn{1}{c|}{2.36} & 12.79 \\
\multicolumn{1}{c|}{} & \multicolumn{1}{c|}{DVQ-VAE} & \multicolumn{1}{c|}{PointNet} & 97.96 & \underline{4.58} & 3.35 & \multicolumn{1}{c|}{0.14} & 2.86 & \multicolumn{1}{c|}{\underline{3.53}} & \underline{3.72} \\
\multicolumn{1}{c|}{} & \multicolumn{1}{c|}{\cellcolor[HTML]{E0E0E0}DVQ-VAE-2} & \multicolumn{1}{c|}{\cellcolor[HTML]{E0E0E0} Mesh UFormer} & \cellcolor[HTML]{E0E0E0} \underline{99.00} & \cellcolor[HTML]{E0E0E0}\textbf{3.35} & \cellcolor[HTML]{E0E0E0}\underline{2.79} & \multicolumn{1}{c|}{\cellcolor[HTML]{E0E0E0}\underline{0.11}} & \cellcolor[HTML]{E0E0E0} 2.82 & \multicolumn{1}{c|}{\cellcolor[HTML]{E0E0E0}\textbf{3.64}} & \cellcolor[HTML]{E0E0E0}\textbf{2.96} \\ \midrule
\multicolumn{1}{c|}{\multirow{5}{*}{\begin{tabular}[c]{@{}c@{}}GRAB\\ ~\cite{taheri2020grab}\end{tabular}}} & \multicolumn{1}{c|}{GraspCVAE~\cite{jiang2021hand}} & \multicolumn{1}{c|}{PointNet} & 97.10 & \underline{3.54} & 2.02 & \multicolumn{1}{c|}{\textbf{0.0041}} & \textbf{2.93} & \multicolumn{1}{c|}{0.81} & 2.48 \\
\multicolumn{1}{c|}{} & \multicolumn{1}{c|}{GraspTTA~\cite{jiang2021hand}} & \multicolumn{1}{c|}{PointNet} & \textbf{100} & 5.05 & \underline{1.74} & \multicolumn{1}{c|}{20.42} & 2.88 & \multicolumn{1}{c|}{0.93} & 2.74 \\
\multicolumn{1}{c|}{} & \multicolumn{1}{c|}{GraspingField~\cite{karunratanakul2020grasping}} & \multicolumn{1}{c|}{PointNet} & 74.80 & 10.56 & 3.80 & \multicolumn{1}{c|}{62.46} & \underline{2.91} & \multicolumn{1}{c|}{3.53} & 5.83 \\
\multicolumn{1}{c|}{} & \multicolumn{1}{c|}{DVQ-VAE} & \multicolumn{1}{c|}{PointNet} & 98.60 & \textbf{3.18} & 2.13 & \multicolumn{1}{c|}{0.15} & 2.83 & \multicolumn{1}{c|}{\underline3.64} & \underline2.45 \\
\multicolumn{1}{c|}{} & \multicolumn{1}{c|}{\cellcolor[HTML]{E0E0E0}DVQ-VAE-2} & \multicolumn{1}{c|}{\cellcolor[HTML]{E0E0E0} Mesh UFormer} & \cellcolor[HTML]{E0E0E0}\underline{99.20} & \cellcolor[HTML]{E0E0E0}3.91 & \cellcolor[HTML]{E0E0E0}\textbf{1.61} & \multicolumn{1}{c|}{\cellcolor[HTML]{E0E0E0}\underline{0.11}} & \cellcolor[HTML]{E0E0E0}2.85 & \multicolumn{1}{c|}{\cellcolor[HTML]{E0E0E0}\textbf{3.86}} & \cellcolor[HTML]{E0E0E0}\textbf{2.30} \\ \midrule
\multicolumn{1}{c|}{\multirow{5}{*}{\begin{tabular}[c]{@{}c@{}}Obman\\ ~\cite{hasson2019learning}\end{tabular}}} & \multicolumn{1}{c|}{GraspCVAE~\cite{jiang2021hand}} & \multicolumn{1}{c|}{PointNet} & 99.20 & 4.32 & \textbf{1.81} & \multicolumn{1}{c|}{\textbf{0.0040}} & \underline{2.95} & \multicolumn{1}{c|}{1.50} & \textbf{2.57} \\
\multicolumn{1}{c|}{} & \multicolumn{1}{c|}{GraspTTA~\cite{jiang2021hand}} & \multicolumn{1}{c|}{PointNet} & \textbf{100} & 5.85 & \underline{2.06} & \multicolumn{1}{c|}{19.70} & \textbf{2.96} & \multicolumn{1}{c|}{1.50} & 3.20 \\
\multicolumn{1}{c|}{} & \multicolumn{1}{c|}{GraspingField~\cite{karunratanakul2020grasping}} & \multicolumn{1}{c|}{PointNet} & 74.62 & 10.53 & 3.81 & \multicolumn{1}{c|}{60.06} & 2.81 & \multicolumn{1}{c|}{2.33} & 5.83 \\
\multicolumn{1}{c|}{} & \multicolumn{1}{c|}{DVQ-VAE} & \multicolumn{1}{c|}{PointNet} & \underline{99.82} & \textbf{3.93} & 2.70 & \multicolumn{1}{c|}{0.14} & 2.90 & \multicolumn{1}{c|}{\underline{3.98}} & 3.07 \\
\multicolumn{1}{c|}{} & \multicolumn{1}{c|}{\cellcolor[HTML]{E0E0E0}DVQ-VAE-2} & \multicolumn{1}{c|}{\cellcolor[HTML]{E0E0E0} Mesh UFormer} & \cellcolor[HTML]{E0E0E0}97.66 & \cellcolor[HTML]{E0E0E0}\underline{4.24} & \cellcolor[HTML]{E0E0E0} 2.31 & \cellcolor[HTML]{E0E0E0}\underline{0.11} & \cellcolor[HTML]{E0E0E0} 2.84 & \cellcolor[HTML]{E0E0E0}\textbf{4.02} & \cellcolor[HTML]{E0E0E0}\underline{2.89} \\ \bottomrule
\end{tabular}

    \label{tab:table_1}
    \vspace{-3mm}
\end{table*}

\subsection{Implementation Details}
\label{subsec:imple}

Our model is implemented using PyTorch on a single NVIDIA RTX 3090 GPU, with a total training time of 1000 minutes. We train our model on the Obman dataset~\cite{hasson2019learning}, sampling \(N_o=3000\) points from the object mesh as input. We adopt the Adam optimizer with an initial learning rate of 1e-4 for 200 epochs, halving the learning rate at epochs 60, 120, 160, and 180. The hyperparameters are set as follows: \(\lambda_e=10\), \(\lambda_m=-50\), \(\lambda_c=1500\), \(\lambda_p=5\), \(\lambda_h=0.1\), and \(\lambda_v=10\). For training PixelCNN~\cite{van2016conditional}, we also employ the Adam optimizer with a learning rate of 3e-4 for 100 epochs. Additionally, the number of layers $L$ of our proposed Mesh UFormer is set to $2$. The voxel size \(\mathcal{V}\) for sampling is set to \(\mathcal{V}=\{0, \frac{O_{size}}{16}, \frac{O_{size}}{8}\}\), where \(O_{size}\) represents the size of the object. Each Mesh UFormer Encoder and Decoder consists of three layers of Point Transformers with normal vector-guided position encoding. The dimensions of \(q\), \(k\), \(v\), and e are \(64\), and the dimension of the hidden layer is \(256\), and the number of neighboring \(K\) is \(16\). We use Adam
optimizer with a learning rate of 1e-5 for 300 epochs to train Mesh UFormer. More details refer to the supplementary material.






\begin{figure*}[thb]
    \centering
    \includegraphics[width=0.95\linewidth]{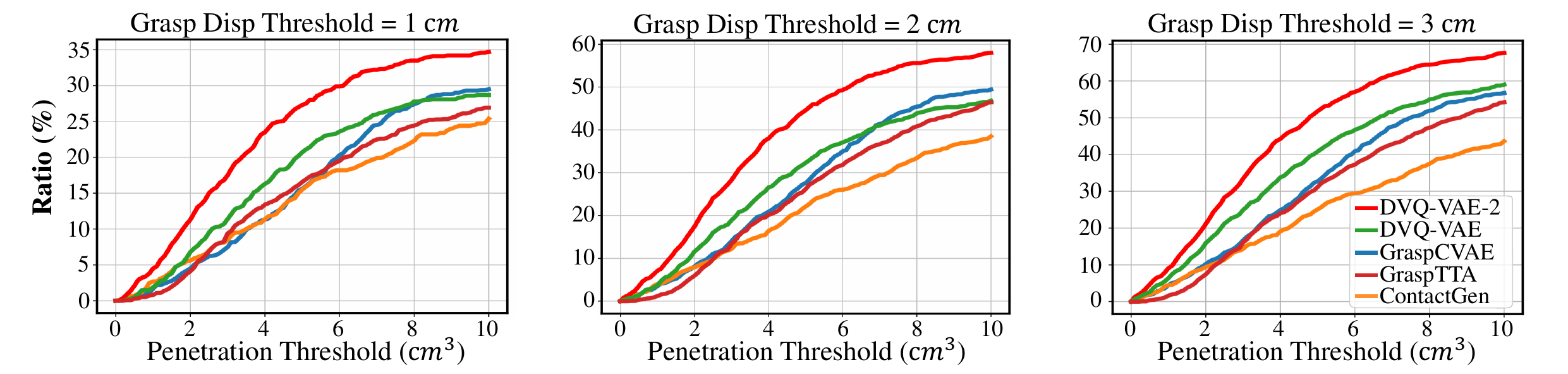}
    \vspace{-3mm}
    \caption{Performance comparison of our proposed method with other models in terms of high-quality ratio, concerning the penetration threshold across different models on the HO-3D dataset.}
    \label{fig:high-qualty}
    \vspace{-3mm}
\end{figure*}

\subsection{Results}
\label{subsec:result}

\begin{table*}[]
\centering
\caption{Performance comparison of our proposed DVQ-VAE-2 and widely used methods in terms of deformable objects on the 
 HMDO dataset~\cite{xie2023hmdo}.}
 \vspace{-3mm}
\begin{tabular}{@{}c|ccccc@{}}
\toprule
Method & \multicolumn{1}{c}{\begin{tabular}[c]{@{}c@{}}Hand-object \\ Contact Distance↓\end{tabular}} & \multicolumn{1}{c}{\begin{tabular}[c]{@{}c@{}}Chamfer \\ Distance↓\end{tabular}} & \multicolumn{1}{c}{\begin{tabular}[c]{@{}c@{}}Laplacian Smoothing \\ Objective↓\end{tabular}} & \multicolumn{1}{c}{\begin{tabular}[c]{@{}c@{}}Contact \\ Ratio \(\%\)↑\end{tabular}} & \begin{tabular}[c]{@{}c@{}}Number of \\ Parameters\end{tabular} \\ \midrule
Originality & 4.32 & 0.00 & 5.68 & 98.15 & - \\ \midrule
PointNet~\cite{qi2017pointnet} & 527.30 & 5787.11 & 40.30 & 64.61 & \textbf{949 K} \\
Deformation Graph~\cite{sumner2007embedded} & 5.60 & 6.61 & 4.12 & 92.07 & \underline{1.8M} \\
PointTransformerV3~\cite{wu2024point} & \underline{5.39} & 4.33 & \underline{3.78} & 89.81 & 91.7 M \\
Object-to-Hand Contact Modeling & 7.09  & 1.48 & 5.16 & 91.84 & 6.3 M \\
Standard Position Encoding & 8.49  & 6.37 & 5.29 & 75.00 & 6.3 M \\
Mesh UFormer Layer Count: 1                        & 7.46                 & 4.61      &5.37       &81.18          & 4.6 M  \\
Mesh UFormer Layer Count: 3                        & 7.74             & \underline{1.37}       &3.79     & \underline{94.44}  &      7.6 M           \\
\rowcolor[HTML]{EFEFEF} Ours DVQ-VAE-2 & \textbf{5.27} & \textbf{1.24} & \textbf{3.40} & \textbf{98.15} & 6.3 M \\ \bottomrule
\end{tabular}

\label{tab:deform}
\vspace{-3mm}
\end{table*}

  \begin{table}[]
  \centering
  \caption{Average scores of grasp generation from human evaluation.}
  \vspace{-3mm}
    
\begin{tabular}{@{}c|c|c@{}}
\toprule
Objects & Methods & Score \\ \midrule
\multirow{4}{*}{\begin{tabular}[c]{@{}c@{}}Rigid \\ Objects\end{tabular}} & DVQ-VAE-2 (w/ Mesh UFormer) & \textbf{3.48} \\
 & DVQ-VAE & {\underline{3.20}} \\
 & ContactGen~\cite{liu2023contactgen} & 3.11 \\
 & GraspTTA~\cite{jiang2021hand} & 3.09 \\ \midrule
\multirow{4}{*}{\begin{tabular}[c]{@{}c@{}}Deformable\\ Objects\end{tabular}} & DVQ-VAE-2 (w/ Mesh UFormer) & \textbf{3.41} \\
 & PointTransformerV3~\cite{wu2024point} & {\underline{3.32}} \\
 & Deformation Graph~\cite{sumner2007embedded} & 2.73 \\
 & PointNet~\cite{qi2017pointnet} & 1.24 \\ \bottomrule
\end{tabular}

    \label{tab:human}
    \vspace{-3mm}
\end{table}

{\bf Quantitative Results.}~We present the performance of our proposed model alongside other state-of-the-art methods in Tab.~\ref{tab:table_1}. All models were trained on the Obman dataset~\cite{hasson2019learning} and tested on HO3D~\cite{hampali2020honnotate}, Grab~\cite{taheri2020grab}, and FPHA~\cite{garcia2018first}. Notably, the objects in HO3D, Grab, and FPHA were not included in the Obman training set and were never encountered during training. As indicated in the table, our DVQ-VAE achieves the lowest penetration and grasp displacement compared to state-of-the-art benchmarks, demonstrating fine-grained control over touch interactions. In terms of the quality index, our model exhibits a remarkable relative improvement of \(13.3\%\) on the HO3D dataset, \(13.7\%\) on FPHA, and \(1.2\%\) on GRAB. We also obtain competitive results for the contact ratio and grasp displacement metrics, which assess grasping stability. Although GraspTTA~\cite{jiang2021hand} shows low grasp displacement, its limited grasp diversity leads to the generation of nearly identical grasps for a given object, as noted in~\cite{liu2023contactgen}. Furthermore, our model achieves a relative improvement of \(22.6\%\) in cluster size, indicating that the part-aware decomposed architecture we incorporated into VQ-VAE~\cite{van2017neural} allows us to match or exceed the performance of leading methods in grasp diversity. We also measured the ratio of high-quality grasps (defined by both penetration and displacement being below certain thresholds). As shown in Fig.~\ref{fig:high-qualty}, we report the ratios of each model on HO-3D objects as the penetration threshold varies from \(0~cm^3\) to \(10~cm^3\), clearly demonstrating that our model outperforms others in most cases. These findings affirm the superiority and efficiency of our DVQ-VAE. After integrating Mesh UFormer, we propose D-VQVAE-2. The results on the rigid object dataset, as shown in Tab.~\ref{tab:table_1}, indicate that Mesh UFormer outperformed PointNet~\cite{qi2017pointnet} in generating higher-quality grasp (w.r.t Quality Index) across all rigid object datasets, achieving a relative improvement of \(5.86\%\) to \(20.43\%\). Additionally, Mesh UFormer demonstrated enhanced grasp diversity with a relative increase of \(0.52\%\) to \(6.04\%\). These improvements are attributed to our mesh-based sampling approach and the position encoding derived from mesh surface normal vectors, enabling the model to capture surface shape information rather than relying solely on object point clouds.

For the deformable objects, we compared our proposed Mesh UFormer with PointTransformerV3~\cite{wu2024point}, PointNet~\cite{qi2017pointnet}, and Deformation Graph~\cite{sumner2007embedded}, and the results are presented in Tab.~\ref{tab:deform}. We can observe that our model generated the most accurate deformations in necessary areas (achieving a \(2.23\%\) improvement in Hand-object Contact Distance compared to PointTransformerV3~\cite{wu2024point}) while minimizing unwanted deformations (a relative 68.82\% improvement in Chamfer Distance over PointTransformerV3~\cite{wu2024point}). Furthermore, it exhibited the least excessive deformations (achieving a \(6.08\%\) absolute improvement in Contact Ratio compared to Deformation Graph~\cite{sumner2007embedded}). Notably, Mesh UFormer achieved these enhanced deformation results with only \(14.56\%\) of the parameters required by PointTransformerV3~\cite{wu2024point}, showing the efficiency of our model.

{\bf Human Evaluation.}~We conducted subjective evaluations by inviting 20 participants to assess the generated grasps from our DVQ-VAE-2, DVQ-VAE, ContactGen~\cite{liu2023contactgen}, and GraspTTA~\cite{jiang2021hand} on rigid objects, as well as subjective assessments of deformations simulated by our DVQ-VAE-2 with Mesh UFormer, PointTransformerV3~\cite{wu2024point}, Deformation Graph~\cite{sumner2007embedded}, and PointNet~\cite{qi2017pointnet} on deformable objects. For testing, participants evaluated 4 randomly generated grasps for each of the eight objects from the HO-3D dataset~\cite{hampali2020honnotate}. They rated the grasps on a scale from 1 to 5 in terms of \emph{penetration depth}, \emph{grasp stability}, and \emph{naturalness}, where higher scores indicate grasps that more closely resemble real human grasps. The results are shown in Tab.~\ref{tab:human}, the subjective evaluation scores of DVQ-VAE showed a relative improvement of \(2.89\%\) compared to previous models. By integrating Mesh UFormer into DVQ-VAE, the subjective evaluation scores of DVQ-VAE-2 demonstrated a relative improvement of \(8.75\%\) compared to DVQ-VAE. For deformable objects, participants rated the physical plausibility of deformations simulated by each model for 20 grasps on a scale of 1 to 5 from HMDO dataset~\cite{xie2023hmdo}. As shown in Tab.~\ref{tab:human}, our model achieved a relative improvement of \(2.71\%\) compared to the best-performing alternative model. 

{\bf Inference Time.}~As shown in Tab.~\ref{tab:table_1}, our proposed DVQ-VAE achieves a time cost reduction of \(99.8\%\) and \(99.3\%\) compared to ContactGen~\cite{liu2023contactgen} and GraspTTA~\cite{jiang2021hand}, respectively, demonstrating a faster generation speed due to the dual-stage decoding. After integrating Mesh UFormer, compared to PointNet, both have a similar number of layers, but the computational complexity of Mesh UFormer decreases progressively with each layer. As a result, the inference speed of DVQ-VAE-2 improved by \(21.43\%\) compared to DVQ-VAE with PointNet.

\begin{figure*}[t!]
    \centering
    \includegraphics[width=0.9\linewidth]{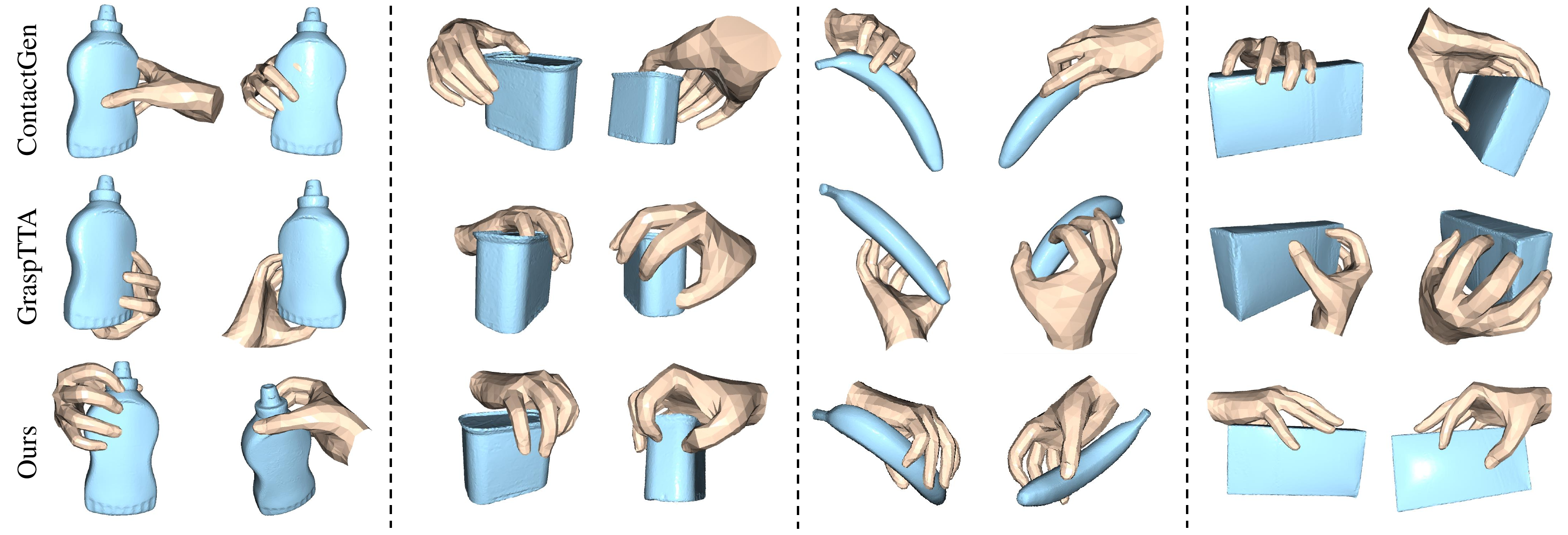}
    \vspace{-3mm}
    \caption{Qualitative results comparing our DVQ-VAE-2 with ContactGen~\cite{liu2023contactgen} and GraspTTA~\cite{jiang2021hand} w.r.t rigid objects on the HO-3D dataset~\cite{hampali2020honnotate}.}
    \label{fig:out-domain}
    \vspace{-3mm}
\end{figure*}

\begin{figure*}[t!]

    \centering
    \includegraphics[width=0.9\linewidth]{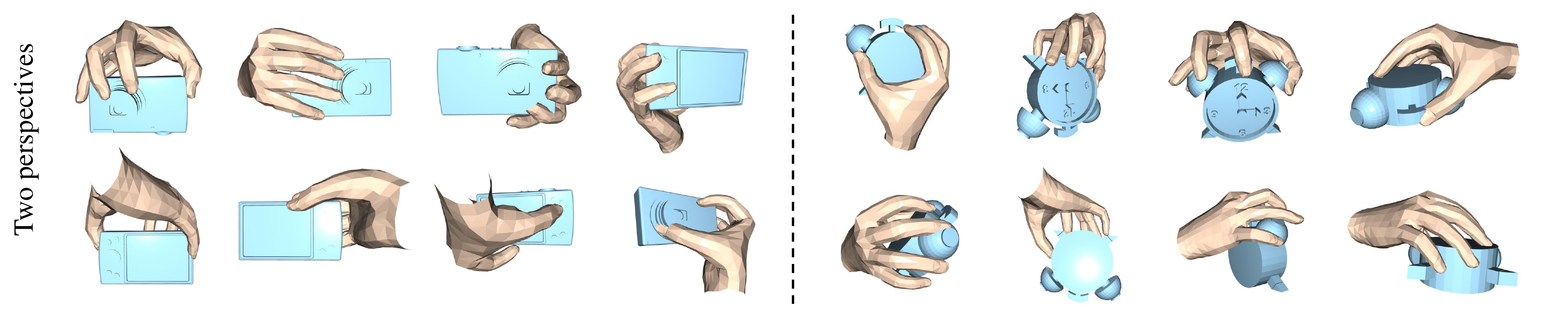}
    \vspace{-5mm}
    \caption{Diverse grasps generated by our proposed DVQ-VAE-2 for the same rigid object in GRAB dataset~\cite{taheri2020grab}.}
    \label{fig:diversity}
\end{figure*}

\begin{figure*}[t!]
    \centering
    \includegraphics[width=0.9\linewidth]{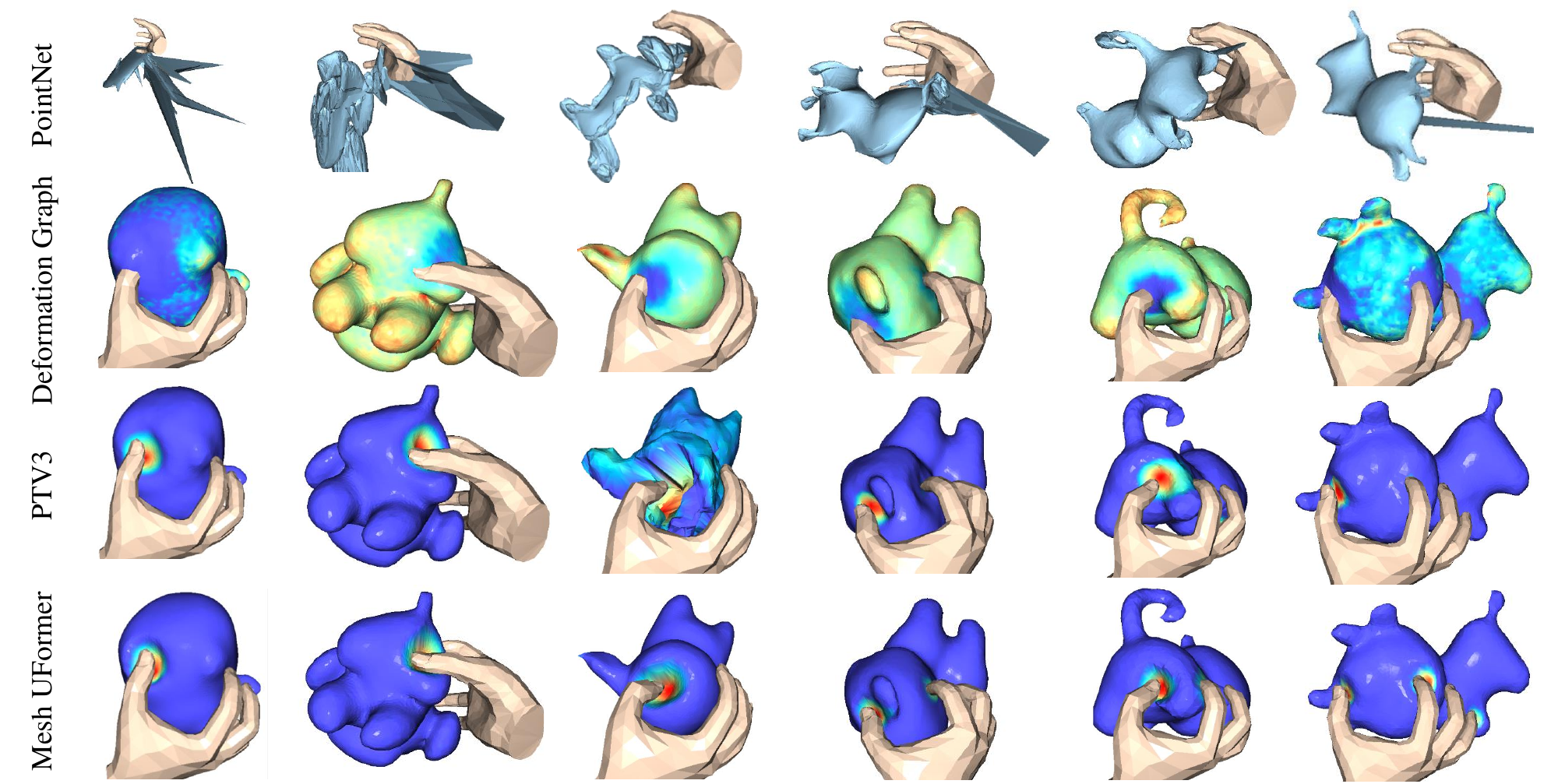}
    \vspace{-3mm}
    \caption{Qualitative results by comparing our proposed DVQ-VAE-2 with Mesh UFormer to Deformation Graph~\cite{sumner2007embedded}, PointNet~\cite{qi2017pointnet}, and PointTransformerV3 (PTV3)~\cite{wu2024point} w.r.t deformable objects on the HMDO dataset~\cite{xie2023hmdo}.}
    \vspace{-3mm}
    \label{fig:softvis}
\end{figure*}

{\bf Qualitative Results.}~We visualize the grasps generated for various rigid objects across different datasets, presenting each grasp from two perspectives. As illustrated in Fig.~\ref{fig:out-domain}, our model consistently generates grasps with fewer penetrations and greater stability than other methods, demonstrating strong adaptability to out-of-domain objects. Moreover, Fig.~\ref{fig:diversity} shows the diverse types of grasps generated for the same object, highlighting our model's ability to produce grasps with a wider variety of postures. 

\begin{figure}[t!]
    \centering
    \includegraphics[width=0.9\linewidth]{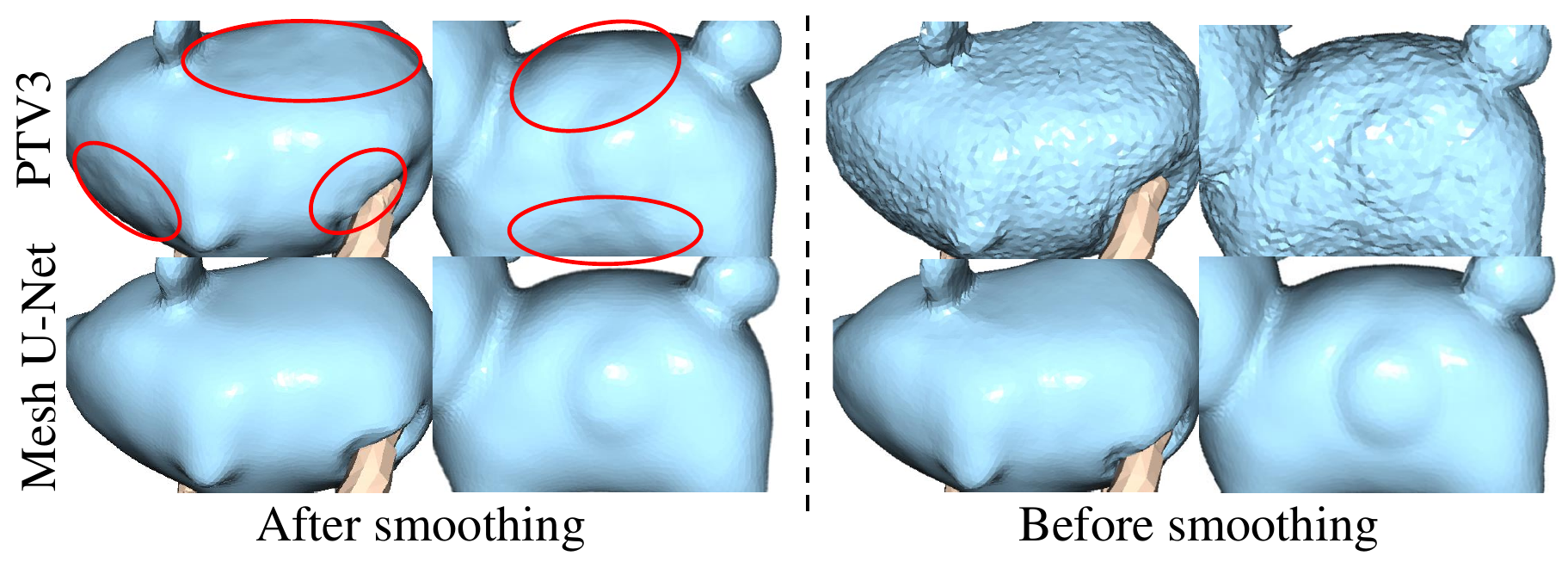}
    \vspace{-5mm}
    \caption{The comparison of deformations simulated by our proposed DVQ-VAE-2 and PointTransformerV3 (PTV3)~\cite{wu2024point}, which are conducted separately before and after smoothing.  It can be observed that the mesh surface within the red circle is not smooth.}
    \vspace{-3mm}
    \label{fig:surface}
\end{figure}

For deformable objects, as illustrated in Fig.~\ref{fig:softvis}, we visualized the deformation results of various models on the HMDO dataset~\cite{xie2023hmdo} and we can clearly observe our method can outperform other approaches. PointNet~\cite{qi2017pointnet}, due to its lack of attention to local information, produced uniform deformations across all points, regardless of whether deformation was needed. Deformation Graph~\cite{sumner2007embedded}, while employing multiple regularization terms to constrain the location and direction of deformations, showed limited effectiveness. This limitation arises because only a minimal number of points initially participate in the deformation process, resulting in less pronounced changes. PointTransformerV3 (PTV3)~\cite{wu2024point} has the capability to understand local object information and involves all points in the deformation simulation. However, experimental results indicate that simulating the deformation of all points simultaneously results in an unsmooth surface as shown in Fig.~\ref{fig:surface}, as PTV3 lacks surface structural information. In contrast, our proposed DVQ-VAE-2 with Mesh UFormer can leverage the hierarchical and multi-resolution grid representations, combined with normal vector-guided position encoding, to comprehensively extract the surface information. This approach enables Mesh UFormer to capture and represent detailed 3D geometric data more effectively, which is essential for accurate deformation simulation. Through a multi-level forward mapping and backward mapping process, the deformations of each point are smoothed, allowing Mesh UFormer to produce more reasonable and natural deformations. More visualized examples and failure cases please refer to our supplementary materials. 


\subsection{Ablation Studies}
\label{subsec:ablat}




\begin{table*}[h]
            \centering
                \caption{Comparison of the performance of our proposed DVQ-VAE and DVQ-VAE-2 and their variants in the ablation study on the HO3D dataset.}
                \vspace{-3mm}
            \begin{tabular}{@{}l|cc|cc|c@{}}
            \toprule
            Method & \begin{tabular}[c]{@{}c@{}}Penetration\\ Volume~\((cm^3)\)↓\end{tabular} & \begin{tabular}[c]{@{}c@{}}Grasp \\ Disp~\((cm)\) ↓\end{tabular} &\begin{tabular}[c]{@{}c@{}} Entropy ↑ \end{tabular}& \begin{tabular}[c]{@{}c@{}}Cluster\\ Size↑\end{tabular} & \begin{tabular}[c]{@{}c@{}}Quality \\ Index~↓\end{tabular}  \\ \midrule
            VQ-VAE  & 6.67 & 7.21 & 2.82 & 1.69 & 7.05 \\
            VQ-VAE-2  &5.18 & 9.87 & \textbf{2.96}& \underline{4.29} & 8.46 \\
            DVQ-VAE  & 10.88 & 4.98 & 2.94 & 4.24 & 6.76\\
            VQ-VAE + Dual-Stage (Two Encoders)  & \underline{4.44} & 3.61 & 2.79 & 1.73& 3.86\\
            DVQ-VAE + Dual-Stage (One Encoder)  & 11.20 & 4.57 & \underline{2.95} & \textbf{4.31} & 6.57\\
            DVQ-VAE + Dual-Stage (Reverse) &7.56 &2.93 & 2.79&2.67 &4.32\\
            DVQ-VAE + Dual-Stage (Two Encoders)  & 5.36 & \underline{2.75} & 2.80 & 3.84 & \underline{3.54}  \\
            DVQ-VAE-2 & 5.30& 3.84 &  2.74&  3.98&  4.28 \\
            DVQ-VAE-2 (One Layer) + Dual-Stage & 5.15& 3.78 &  2.80&  3.96&  4.20 \\
            DVQ-VAE-2 (Three Layers) + Dual-Stage & 4.94& 3.51 &  2.87&  3.37&  3.94 \\
            DVQ-VAE-2 (Two Layers) + Dual-Stage  &\textbf{4.24}  & \textbf{2.31} &2.84  &4.02  &\textbf{2.89} \\  
            \bottomrule
            \end{tabular}

    \label{tab:table_ablation}
\end{table*}



\textbf{Object Encoder} We compare the results of encoding the given object with only one object encoder and with our proposed two object encoders in Tab.~\ref{tab:table_ablation}. We can find our proposed DVQ-VAE with two encoders outperforms the model with only one encoder across most of the metrics, which is attributed to the two encoders that can empower the object representation by decoupling the object's feature into type and pose parts.


\textbf{Part-aware Decomposed Architecture.} We ablate the decomposed architecture and present the performance in Tab.~\ref{tab:table_ablation}. The proposed architecture demonstrates a relative improvement of \(22.4\%\) in grasp stability and \(151\%\) in cluster size compared to the vanilla VQ-VAE~\cite{van2017neural}. This indicates that our part-aware decomposed architecture enhances both the diversity and quality of generated grasps. In terms of the backbone network selection, we also evaluated the VQ-VAE-2~\cite{razavi2019generating} structure, which incorporates global features. However, as indicated in Tab.~\ref{tab:table_ablation}, the results were not satisfactory. In contrast, the autoregressive model PixelCNN~\cite{van2016conditional} effectively captures latent or implicit global hand features during reconstruction.

\begin{figure}[t!]
    \centering
    \includegraphics[width=0.9\linewidth]{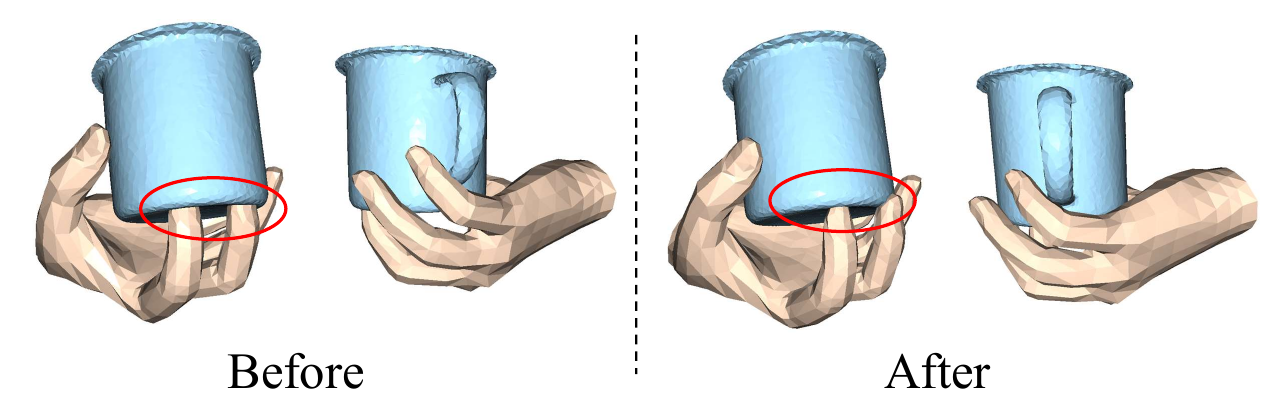}
    \vspace{-5mm}
    \caption{Comparison of generated grasps with and without our dual-stage decoding strategy, labeled as ``after'' and ``before'', respectively.}
    \vspace{-3mm}
    \label{fig:refine}
\end{figure}

\textbf{Dual-Stage Decoding Strategy.} We evaluate the effectiveness of our proposed dual-stage decoding strategy in Tab.~\ref{tab:table_ablation}. Compared to models without this strategy, our full model demonstrates an increased contact ratio, and reduced penetration and displacement, leading to improved overall grasping quality. Specifically, this strategy results in \(45.2\%\) and \(47.6\%\) relative increases in the quality index for our DVQ-VAE and VQ-VAE~\cite{van2017neural}, respectively. Additionally, Dual-Stage Decoding Strategy improved the Quality Index of DVQ-VAE 2.0 by \(32.48\%\). We also explored decoding position before posture, but the "Dual-Stage (Reverse)" results in the table were not promising. Furthermore, the position generation module in our trained DVQ-VAE optimizes grasps generated by other models by refining the grasping positions. As shown in Fig.~\ref{fig:refine}, the optimized grasps exhibit reduced penetration, confirming the effectiveness of our dual-stage decoding strategy.

\textbf{Mesh UFormer.} For deformable objects, as shown in Tab.~\ref{tab:deform}, our proposed method for quantifying hand-object contact demonstrates significant advantages over traditional Object-to-Hand Contact Modeling approaches commonly used for rigid objects. As discussed in Sec.~\ref{HoC}, traditional methods fail to accurately represent the true penetration distance of the hand into the object, leading to an increase of \(34.54\%\) in the Hand-Object Contact Distance metric, which evaluates deformation accuracy. We also tested the performance of Mesh UFormer with varying numbers of layers. Results indicate that deformation accuracy improves initially with an increase in layers but declines beyond a certain point. The optimal performance was achieved with a two-layer Mesh UFormer. This aligns with the phenomenon where stacking too many graph convolutional layers results in the loss of deep local features~\cite{li2018deeper}. Excessive layers lead to the loss of detailed local surface information, which, despite improving the smoothness of generated deformations and reducing the Laplacian Smoothing Objective by \(10.48\%\), significantly decreasing deformation accuracy and increasing the Hand-Object Contact Distance by \(46.87\%\). Conversely, using too few layers results in insufficient forward mapping and backward mapping processes for each point’s features, leading to less smooth deformations and a \(57.94\%\) increase in the Laplacian Smoothing Objective. For rigid objects, as shown in Tab.~\ref{tab:table_ablation}, for the same reasons as mentioned above, the three-layer Mesh UFormer and the two-layer Mesh UFormer increased the Quality Index by \(26.65\%\) and \(45.33\%\), respectively.

\textbf{Deformation Modeling.} For grasp generation of deformable objects, we proposed a novel normal vector-guided positional encoding approach. In contrast to conventional position-based encoding, which only considers the Euclidean distance between points, our approach integrates local geometric context by utilizing normal vectors, which enhances the model's ability to capture the deformation characteristics of the object. As demonstrated in Tab.~\ref{tab:deform}, the normal vector-guided encoding significantly improved the model’s performance in grasping deformable objects, yielding a remarkable \(37.93\%\) reduction in Hand-Object Contact Distance compared to the standard positional encoding method. This improvement highlights the importance of incorporating local surface geometry for more accurate deformation modeling during grasping.

\section{Conclusion}
\label{sec:conclu}

In this paper, we presented a novel generative model, \ie, Decomposed VQ-VAE, for human grasp generation with rigid or deformable objects. The proposed part-aware decomposed architecture can blend several latent codebooks to account for different components of the hand, and the dual-stage decoding strategy can make the hand posture fit into the position in order. Furthermore, for deformable objects, Mesh UFormer was newly-designed as the backbone to enhance the hierarchical structure of the object mesh. Meanwhile, we showed a normal vector-guided position encoding approach for deformation modeling. We have shown that each one of these components was important and helped outperform state-of-the-art methods. In the future, we will apply our model to human-robot interaction and embodied AI.

\ifCLASSOPTIONcaptionsoff
  \newpage
\fi



%

\bibliographystyle{IEEEtran}
\bibliography{main}

\end{document}